\documentclass[acmlarge, screen,nonacm]{acmart}
\setcopyright{none}
\renewcommand\footnotetextcopyrightpermission[1]{}
\usepackage{booktabs}
\usepackage{subfig}
\usepackage{multirow}

\AtBeginDocument{%
  \providecommand\BibTeX{{%
    Bib\TeX}}}

\usepackage{mathtools}
\usepackage{textgreek}
\def\BibTeX{{\rm B\kern-.05em{\sc i\kern-.025em b}\kern-.08em
    T\kern-.1667em\lower.7ex\hbox{E}\kern-.125emX}}

\begin{document}

\title{An Effective, Reliable, and Robust Framework for Human Activity Recognition Using Wearable Sensors}

\author{Nafees Ahmad}
\authornote{This is the corresponding author.}
\affiliation{%
  \department{Department of Computer Science and Engineering}
  \institution{The Chinese University of Hong Kong}
  \city{Hong Kong SAR}
  \country{China}
}
\email{nafees@link.cuhk.edu.hk}

\author{Ho-fung Leung}
\affiliation{%
  \institution{Independent Researcher}
  \city{Hong Kong SAR}
  \country{China}}
\email{ho-fung.leung@outlook.com}

\author{Muhammad Adil Abid}
\affiliation{%
  \department{Department of Computer Science and Media Technology}
  \institution{Malm{\"o} University}
  \city{Malm{\"o}}
  \country{Sweden}
}
\email{muhammad.adil-abid@mau.se}

\author{Sadia Shakil}

\affiliation{%
  \institution{National University of Sciences and Technology (NUST)}
  \city{44000, Islamabad}
  \country{Pakistan}
}

\affiliation{%
  \department{Department of Biomedical Engineering}
  \institution{The Chinese University of Hong Kong}
  \city{Hong Kong SAR}
  \country{China}
}
\email{sadia.shakil@seecs.edu.pk}
\email{sadiashakil@cuhk.edu.hk}

\renewcommand{\shortauthors}{Nafees et al.}

\begin{abstract}
Human Activity Recognition (HAR) through wearable sensors greatly improves the quality of human life through its multiple applications in health monitoring, assisted living, and fitness tracking. For HAR, multi-sensor channel information is vital for optimal performance. Current work states that applying an attention neural network to prioritize discriminatory sensor channels helps the model classify activity more precisely. However, obtaining discriminatory information from multisensory channels is not always trivial, for example, when collecting data from older hospitalized patients. In this context, existing HAR methods struggle to classify activities, particularly activities with similar natures. Moreover, HAR deep models predominantly suffer from overfitting due to the small size of available datasets, which leads to poor performance. Data augmentation is a viable solution to this problem. However, currently available data augmentation methods to HAR have various drawbacks, including the possibility of being domain-dependent, resulting in distorted models for test sequences. To address these HAR problems, we propose a novel framework, ALAE-TAE-CutMix\textsuperscript{+}, which primarily focuses on two aspects. First, it enhances the latent information across each sensor channel and learns to exploit the relation among multiple latent features and the ongoing activity. Consequently, the discriminatory feature representations of each activity is enriched. Second, a new augmentation strategy is introduced to address the shortcomings of existing multi-sensor channel data augmentation. We then extend the framework to create a further enhanced version, namely ALAE-CIE-TAE-CutMix\textsuperscript{+}, which learns to capture the interactions between the features of each pair of sensor channels. We find that although the first framework performs slightly better than the latter, the latter is nonetheless more reliable and robust. Both frameworks significantly outperform existing state-of-the-art approaches on the four most commonly used HAR datasets from diverse domains. In addition, we also validate the contribution of each of the modules in the proposed frameworks through detailed ablation studies.
\end{abstract}





\maketitle

\section{Introduction} 
Wearable sensor-based human activity recognition (HAR) has gained much popularity due to the proliferation of low-cost, infrastructure-less, comfortable-to-wear sensing devices. It provides a wide range of applications, particularly in the healthcare domain, for examples, patient monitoring, assistance for the aged, and rehabilitation~\cite{9, 10, 12, 13, 15, 33, 40, ahmad2019sarm}. A key element in building these applications is to have a model capable of automatically and accurately recognizing human activity using wearable sensors.

Deep learning (DL) based models have established benchmark performance in HAR~\cite{s2016, satt2018, sCNNLSTM2016, sattend2021}, and the performance of these models is highly dependent on multi-sensor channels data. Some sensor channels provide more discriminatory information for ongoing activity than others, and giving attention to these sensor channels helps the model classify the activities more accurately~\cite{sattend2021}. However, in some situations, receiving discriminatory information from most sensor channels is non-trivial, particularly when capturing older adults' similar yet different activities. More specifically, for subjects who are hospitalized and older, they cannot perform activities in the same manner as younger and healthier people because they cannot actively twitch their body muscles due to illness, frailty, and aging. In addition, older people can remain stationary for prolonged periods and often they cannot move, especially in activities of static nature. In such situations, the attached sensors cannot capture signals from the human body in the same way as they can from a healthy young body. As a result, 
very minor fluctuations (even almost negligible) in activity patterns may yield information that makes it more difficult to classify the activity. Encountering such a situation, where activities of similar nature  are also present, it becomes more 
challenging to obtain discriminatory information about these activities. This results in a higher likelihood that these activities will be confused with one another during the classification process. Under these circumstances, getting the sensor channel’s attention does not help too much the model classify the activities.

Available HAR datasets based on wearable sensors are typically small due to the tedious effort required to collect the labeled data~\cite{sur2012, DACNN2018}. Limited training data could lead to model overfitting and unsatisfactory results on unseen test data. Augmenting the training data with artificially generated sequences contributes to generalizing model performance across many other domains such as computer vision. However, current HAR studies are largely limited to handcrafting methods (scaling, jittering, {\it etc.}), assuming they produce virtual sequences by altering original sequences while preserving their label semantics. In these generated sequences, the semantics and salient features of the original label data may not be preserved, thus generating irrelevant and unrecognizable sequences that may potentially mislead the model during training~\cite{41}. Abedin, Ehsanpour, Shi, Rezatofighi and Ranasinghe~\cite{sattend2021} use the {\it mixup} data augmentation strategy in HAR, where two original sequences are linearly mixed. Although this improves the model's performance to some extent, the augmented mixup sequences may look completely different from the original sequences, causing the mixed sequence to become meaningless from a human perspective. Also, linear mixing can mix the salient features of the two original sequences, leading to a completely different representation that cannot associate with any of the original sequences. Therefore, training on these generated sequences may distort the learned model for the test sequences.

This paper is extension of our previous work~\cite{ahmad2023alae}. We propose a HAR framework called ALAE-TAE-CutMix\textsuperscript{+}, comprised of multiple new components to address the aforementioned problems. To address the lack of discriminatory information, we propose an adaptive latent attention encoder (ALAE) to enhance latent information across each sensor channel and exploit the relationship between multiple latent information of each sensor channel and corresponding activity. The ALAE only transmits those representations to subsequent modules that can better encode the current activity. Furthermore, Long Short Term Memory (LSTM) with a temporal attention layer can enhance performance by learning the temporal and contextual relationships in the sensor sequence~\cite{satt2018}. Therefore, we adopt a temporal attention encoder (TAE) to improve the generated representations of ALAE by further learning the activity's temporal and relevant information.

To prevent the augmented sequence from large alteration and perform a meaningful adoption, we propose to adopt a temporal CutMix method to regularize our model. Unlike the mixup method~\cite{sattend2021}, it simply copies the subsequence region of one sequence into another sequence to generate the virtual sequence. The augmented sequence label is generated based on the ratio of both original sequence contributions, effectively resolving the issue of label-preserving semantics raised in handcrafted
augment sequences. Additionally, the CutMix technique is easy to apply and it is dataset independent. It penalizes the deep model for learning non-discriminatory regions. Experimental results suggest a significant generalization improvement via the proposed temporal CutMix augmentation.

However, in some cases, the virtual sequence is generated from entirely different properties of two activity sequences, such as ``cycling'' and ``sleeping,'' which can appear unnatural and unrecognizable to the human eye and may hinder the performance of the DL model. Given this, we decide to broaden our investigation of data augmentation for HAR by extending our proposed CutMix method to CutMix\textsuperscript{+}, according to which
the model is trained using a combination of virtual and original sequences. More specifically, we overcome the problem by training the model with the joint supervision of the original sequences loss
(cross-entropy loss) and the virtual sequences loss (CutMix loss) in an end-to-end learning manner.

Another significant challenge to address is learning the possible interactions between the sensor channels. 
As the interactions between sensor channels vary from activity to activity, capturing these interactions might be able to enhance the performance of the model. For example, when cleaning the table, there appears to be a special relationship between the movement of the right hand and the back, indicating notable interactions between these two body positions. Capturing these relations through attached sensors might help to recognize the cleaning table activity more accurately. In fact, the study proposed by Abedin et al.~\cite{sattend2021} shows that learning the interactions between sensor channels can improve the model's performance. Therefore, we hypothesize that exploring the interactions between sensor channels can significantly improve the model performance. To this end, we extend our ALAE-TAE-CutMix\textsuperscript{+} framework with a new component Cross-Channel Interaction Encoder (CIE) referred to as the ALAE-CIE-TAE-CutMix\textsuperscript{+} framework in order to further learn interactions between the sensor channels.

Both our proposed methods achieve superior performance scores across multiple challenging datasets containing data on older hospitalized patients, healthy subjects over the age of ``$50$'', and daily static and sporadic human activities. 
The main contributions of this work are summarized as follows:
\begin{enumerate}
    \item We propose a HAR framework called ALAE-TAE-CutMix\textsuperscript{+}. Where the ALAE module is for enhancing sensor channel information, the TAE module learns temporal contextual information. Moreover,  CutMix augmentation is for HAR model regularization, while CutMix\textsuperscript{+} is an extension of CutMix to address the ambiguity that arises in CutMix augmented sequences in some scenarios.  
    \item We extend the ALAE-TAE-CutMix\textsuperscript{+} framework by integrating the CIE module to form a more comprehensive framework referred to as ALAE-CIE-TAE-CutMix\textsuperscript{+} to take into account the interactions between each pair of sensor channels.
    \item Both our frameworks significantly outperform the state-of-the-art HAR frameworks on four benchmark datasets, thus highlighting the effectiveness, reliability, robustness, and generalization ability of the proposed frameworks. Moreover, through extensive ablation studies, we present the individual contributions made by each module in the proposed frameworks. Furthermore,  we show further insights of our frameworks  by conducting experiments on limited amounts of training data and different segment sizes.  
\end{enumerate}

The paper is structured in the following manner: Section~\ref{sec:RW} briefly reviews the related work on the learning frameworks and data argumentation methods for wearable sensor-based HAR. In Sections~\ref{sec:ALAE} and~\ref{sec:EXT}, respectively, the ALAE-TAE-CutMix\textsuperscript{+} and ALAE-CIE-TAE-CutMix\textsuperscript{+} frameworks' designs and workflows are described. Section~\ref{sec:results} presents the experimental analysis and results of our frameworks, and finally, Section~\ref{sec:CON} ends with the paper's conclusion.

\section{Related Work} \label{sec:RW}
Traditional sensor-based human activity recognition (HAR) process requires segmentation of input data streams using sliding windows, followed by manual feature extraction and finally machine learning classification~\cite{sur2014,shoaib2014fusion, ehatisham2020opportunistic}. These handcrafted techniques achieve somewhat satisfactory results but rely heavily on trial and error and human experience. In order to eliminate the need for time-consuming human feature engineering, deep learning (DL) models are used to extract features automatically~\cite{wang2019deep}. Preliminary studies explore convolutional neural network (CNN) to automatically perform feature extraction in end-to-end learning models, and they are shown to outperform traditional handcrafted machine learning techniques~\cite{CNN1, CNN2, CNN3, CNN4}. These studies use 1D convolution kernels along the time dimension to capture the dependency of consecutive locally connected data samples and different activity pattern variations, thus obtaining salient features in a more abstract representation format. Although these models achieves higher performance, they are not good enough for learning the temporal relations in long sequences. Meanwhile, the HAR community adopts deep recurrent neural networks (RNNs) based methods, for example, forward and bidirectional long short-term memory (LSTM)~\cite{s2016} and various ensemble methods of LSTM models~\cite{LSTM}, to alleviate the problem in capturing long temporal relationships between consecutive sensor input time-stamps. To exploit the strengths of CNN and LSTM, Ord\'{o}\~{n}ez and Roggen~\cite{sCNNLSTM2016} propose a hierarchical combination of CNN and LSTM called DeepConvLSTM, which learns temporal correlations in samples on a more abstract level. The RNN-based models surpass all the previous HAR models in terms of performance. However, they take into consideration all input time-stamps in learning, which may include irrelevant or noisy time-stamps along with the relevant time-stamps of the activity. Murahari and Pl\"{o}tz~\cite{satt2018} mitigate this issue by extending DeepConvLSTM, adding a temporal attention layer to learn relevant time-stamps within the input sequence relevant to the ongoing activity.

More recently, Abedin~{\it et~al.}~\cite{sattend2021} investigate the role of different sensor channels in capturing human activities. They highlight that each sensor channel has distinct sensing capabilities for capturing the same activity. Therefore, they propose a self-attention based deep model to give more attention to those sensor channels that are more important to the current in-process activity. Although the sensor channel attention improves HAR model performance, it may not yield discriminatory information in some situations, for example, in the case of recognizing activities of older hospitalized patients. 

On the other hand, though data augmentation has gained wide popularity in different machine learning domains due to its great ability to regularize DL models, however, relatively less efforts have been made to explore data augmentation methods for multichannel sensor data in the HAR domain. In this context, different  augmentation methods are explored~\cite{41, SOL, SelfHAR}, such as
noise addition, scaling, rotation, inversion, time reversion, signal section scrambling, time stretching and warping, and channel shuffling. Faridee, Khan, Pathak and Roy~\cite{11} explore similar augmentation methods to train the semi-supervised transfer learning-based model for recognizing complex human activities. While these methods show a high potential to generalize the model for unseen test data, they are dataset dependent and require a high level of expertise to be adopted effectively. 
This creates a number of challenges and issues when applying these techniques to HAR multichannel data. In a further development in this area, Abedin~{\it et~al.}~\cite{sattend2021} examine the {\it mixup} data augmentation for HAR, which linearly mixes two examples and their labels (according to the proportion of each example) in each batch to generate a virtual example in training. 
However, due to the linear mixing of two original sequences, the augmented sequence can appear completely different from the original sequences and becoming meaningless from a human perspective, which can affect the performance of the model. 


\section{The ALAE-TAE-CutMix\textsuperscript{+} Framework} \label{sec:ALAE}
In this Section, we describe an end-to-end HAR model ALAE-TAE-CutMix\textsuperscript{+}, which receives human activity data from wearable sensing devices as input and recognizes the ongoing human activities~\cite{ahmad2023alae}. In general, a wearable sensing device used in human activity recognition is composed of multiple sensors of different modalities, such as accelerometers, gyroscopes, magnetometers, and so on. Each of these sensors measures different aspects of motion and orientation, and outputs the measurements from a number of sensor channels ({\it e.g.}, $x$-, $y$-, and $z$-axes).

\begin{figure}[] 
  \centering
  \includegraphics[width=\linewidth]{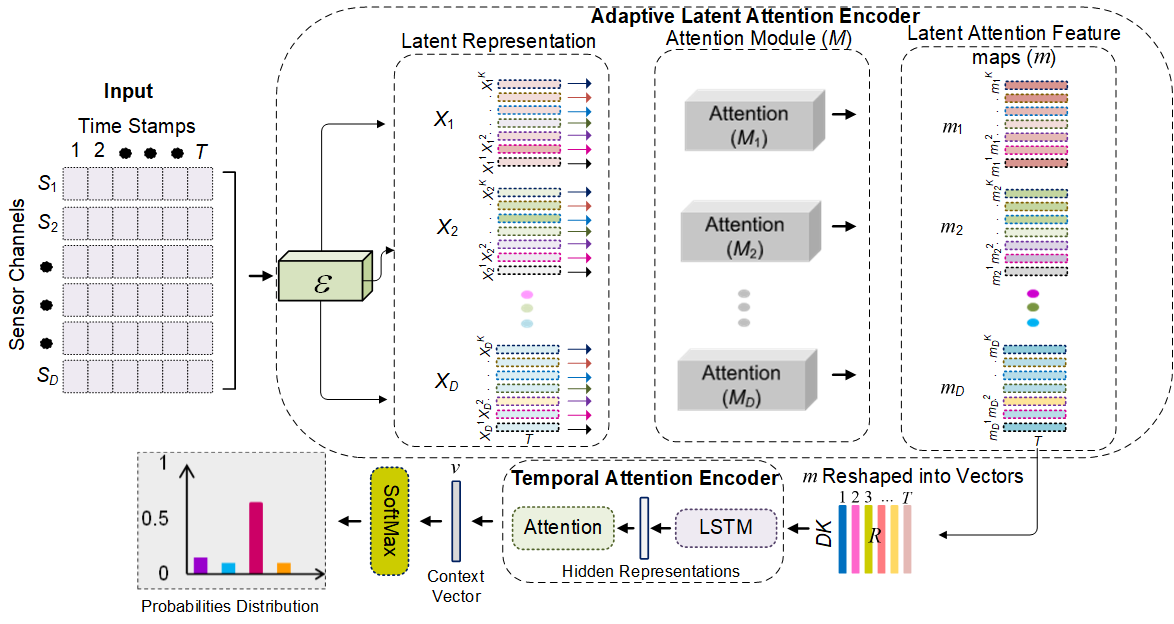}
  \caption{Overview of the proposed ALAE-TAE-CutMix\textsuperscript{+} framework for human activity recognition.}
  \label{fig:FW}
\end{figure}

As depicted in Figure~\ref{fig:FW}, the ALAE-TAE-CutMix\textsuperscript{+} architecture consists in two major modules: the Adaptive Latent Attention Encoder (ALAE) and the Temporal Attention Encoder (TAE). ALAE consists in two components: the encoder \text{\textepsilon} and the attention module $M$. The encoder module \text{\textepsilon} generates the latent representations for each of the $D$ sensor channels using a convolutional neural network. There are $K$ convolutional kernels used to generate $K$ latent representations for each of the sensor channels. Then, the attention module processes these latent representations and produces latent attention feature maps to uncover those sensor channel representations that encode ongoing activity better than other latent representations. Afterwards, TAE learns the sequential relationships between time-stamps in a refined representation of all sensor channels and generates a contextual summary of the relevant time-stamps for the final classification.

Another important component of ALAE-TAE-CutMix\textsuperscript{+} is the CutMix\textsuperscript{+} data augmentation scheme.
Available HAR benchmark datasets, such as Hospital~\cite{sdense2018}, Skoda~\cite{skoda},~{\it etc.}, are typically small in size due to the arduous procedure of gathering annotated data through wearable devices. 
Any DL model becomes prone to overfitting and loses their capacity to generalize. CutMix\textsuperscript{+}, as we shall see, significantly contributes to the improvements of the performance of the model by augmenting the datasets with  segments artificially generated using existing segments.

\subsection{Adaptive Latent Attention Encoder (ALAE)}
As pointed out by Abedin~{\it et al.}~\cite{sattend2021}, in human activity recognition, each  channel of a sensor in a sensing device has a unique perceptual ability to capture the same activity. Some sensor channels provide more discriminatory information for ongoing activity compared to others. Consequently, giving high priority to data from more discriminating sensor channels can improve the performance of the HAR model. However, this is not always true in every situation. When data is acquired from senior individuals and hospitalized patients, most sensor channels provide non-discriminatory information across similar activities ({\it e.g.}, sitting and sitting-down, lying and lying-down, {\it etc.}). For example, it can be seen in the very same study by Abedin~{\it et al.}~\cite{sattend2021} that the proposed model misclassifies most of the sitting-down activity test samples of hospitalized patients as sitting and the lying-down activity samples as lying.  Moreover, when the similar activities (such as lying, sitting, sitting on the sofa, sitting on the chair, sitting on the couch, {\it etc.}) are static in nature, the situation becomes even worse. As aged patients are generally unable to perform activities in the same manner as young, healthy people, they remain in a single position for an extended period of time without many movements, resulting in activity patterns with few spikes or changes. Hence, it is likely that sensor channels do not convey discriminatory information for similar activities of static nature. Therefore, in these cases, encoding the sensor channel with attention is insufficient to improve HAR model performance.

The main task of Adaptive Latent Attention Encoder (ALAE) is to generate multiple different representations for each sensor channel. We hypothesize that the emergence of multiple different latent representations enhances the likelihood of obtaining discriminating information. Then, we can leverage latent representations of those sensor channels that are more informative about the ongoing activity than the representations of other sensor channels. These latent representations can assist the model in encoding activities of similar nature in latent space differently from one another. Accordingly, we design an end-to-end trainable ALAE module that accepts sensor channel representations as input, generates multiple latent representations and produces the attentive latent representations for each sensor channel.
\subsubsection{Latent Representation Generation of Sensor Channels}
To project the input representation into the latent representation, we adopt a trainable encoder \text{\textepsilon} based on a convolutional neural network, which accepts a segment of $T$ time-stamps of input data $S_i\in \mathbb{R}^T$ from sensor channel $i$, $1\leq i \leq D$,  as input and generates $K$ latent representations $X_i^1,\ldots,X_i^K$ corresponding to the sensor channel $i$,
\begin{equation}
    \begin{split}
        \text{\textepsilon}(S) & = \text{\textepsilon}(S_1, S_2,\cdots, S_D) \\
       & = ((X_1^1, X_1^2, \cdots, X_1^K), (X_2^1, X_2^2, \cdots, X_2^K), \cdots, (X_D^1, X_D^2, \cdots, X_D^K)) \\
       & = (X_1, X_2,\cdots, X_D),
    \end{split}
\end{equation}
where $X_i^k \in \mathbb{R}^T$ denotes the $k^\text{th}$ latent representation generated across sensor channel $i$,  $X_i \in \mathbb{R}^{K \times T}$  is the latent space feature maps of sensor channel $i$. 
The convolutional encoder produces these representations, which are subsequently fed into the attention module to generate attentively optimized representations for each sensor channel.

\begin{figure}[] 
  \centering
  \includegraphics[width=\linewidth]{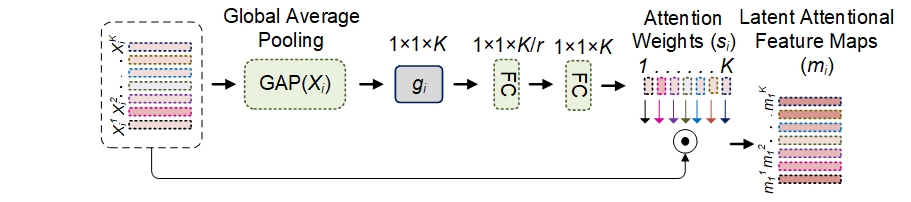}
  \caption{Overview of the attention module.}
  \label{fig:at} 
\end{figure}

\subsubsection{Attention Module (M)}
Inspired by the squeeze and excitation method~\cite{SE}, we design the attention module $M$, where attention network $M_i$ accepts the $K$ latent representations of the $i^\text{th}$ sensor channel and learns the important latent representations among them. Figure~\ref{fig:at} shows the internal structure of attention module $M$. First, we squeeze the data by computing the global average pooling across $K$ latent representations of a sensor channel $i$,
\begin{equation}
    g_i = GAP(X_i) =(g_i^1, g_i^2, \cdots, g_i^K),
\label{eq-1}
\end{equation}
where $g_i \in \mathbb{R}^k$ is a vector of $K$ values generated for each sensor channel $i$. The $k^\text{th}$ value $g_i^k$ is derived from the corresponding $k^\text{th}$ latent representation $X_i^k$. $GAP$ denotes the global average pooling operation applied on latent representations $X_i$ generated by encoder \text{\textepsilon}.

To use the $g_i$ vector produced by the squeeze operation (Equation~\ref{eq-1}), we perform
an excitation operation to enhance the latent representations.  The excitation operation performs two operations to accomplish this task. The first step is to compute attention weights, and the second step is to assign these weights to the corresponding sensor channel latent representations. To obtain the attention weights, all the $K$ values of $g_i$ for each latent representation $X_i$ are passed through a two-layer fully connected (FC) neural network,
\begin{equation}
    \begin{split}
        s_i & = \sigma(h(g_i,W))=\sigma(W_2 \delta(W_1 g_i)) \\
        & = (s_i^1,s_i^2,...,s_i^K),
    \end{split}
\end{equation}
where $\delta$ is the ReLU activation function, $W_1 \in \mathbb{R}^{\frac{K}{r}\times K}$ and $W_2 \in \mathbb{R}^{K\times \frac{K}{r}}$ are the learning parameters associated with
the first and second layers of the FC. To reduce computational complexity and the likelihood of overfitting, and increase
generalization, the number of learning parameters  $W_1$ are reduced by some reduction ratio $r$. However, in the second layer, we increase the dimension again with learning parameters $W_2$  and pass the result through the sigmoid function $\sigma$ to obtain the exact number of attention weights (one for each latent representation). The generated weights $s_i \in \mathbb{R}^K$ are multiplied with the original sensor channel  latent representations $X_i\in \mathbb{R}^T$  respectively to obtain the final enhanced latent representations,
\begin{equation}
    \begin{split}
    m_i &  = s_i X_i \\
    &  = (s_i^1 X_i^1, s_i^2 X_i^2, \cdots, s_i^K X_i^K) \\
    & =(m_i^1, m_i^2, \cdots, m_i^K) 
    \end{split}
\end{equation}
where $m_i^k =((m_i^k)_1, (m_i^k)_2, \cdots, (m_i^k)_t) \in \mathbb{R}^{T}$ indicates the $k^\text{th}$ latent representation across sensor channel $i$, and $m_i =(m_i^1, m_i^2, \cdots, m_i^K) \in \mathbb{R}^{K \times T}$ represents the $K$ number of latent attentive representations of $X_i=(X_i^1, X_i^2, \cdots, X_i^K)$. Overall, from the result of attention module $M$, we get 
$m=(m_1, m_2, \cdots, m_D)\in \mathbb{R}^{D \times K \times T}$, which is passed to the subsequent module of the model.

\subsection{Temporal Attention Encoder (TAE)}\label{TAE}
The resultant feature maps $m=(m_1, m_2, \cdots, m_D)\in \mathbb{R}^{D \times K \times T}$ produced by ALAE are sequentially provided to a two-layer LSTM in order to incorporate a contextual representation of the time-stamps into the model. 
We reshape these feature maps at each time-stamp $t$, $1\leq t\leq T$ to obtain a vector representation $(R_t)_{t=1}^{T}$ for sequential learning, where $R_t = ((m_1^1)_t,\dots, (m_1^K)_t, (m_2^1)_t,\dots, (m_2^K)_t,\dots, \dots,(m_D^1)_t,\dots, (m_D^K)_t)\in \mathbb{R}^{DK}$. 

When considering only the LSTM, we obtain a summarized version of all time-stamps from the last LSTM cell. However, in HAR,  data at different time-stamps do not all contribute equally to recognizing the ongoing activity. Therefore, to alleviate this problem and to reduce the burden from the last hidden state cell of the LSTM, we employ a temporal attention approach after the LSTM final layer,
\begin{equation}
    \alpha_t = \frac{\exp(\tanh(Wh_t))}{\sum_{t=1}^{T}\exp(\tanh(Wh_t))},
\end{equation}
Here, we do not only take the last cell vector representation $h_{T}$ at time-stamp $T$ from the LSTM but also consider all hidden state representations $(h_t)_{t=1}^{T}$ corresponds to each time-stamp $t$, $1\leq t\leq T$. This helps to review all time-stamps and put weights on the more relevant ones. 
In the above formulation, hidden states $(h_t)_{t=1}^{T}$ pass through the first layer to get attention scores, where $W$ is the learning parameter and $\tanh$ is the activation function. Then, the softmax function is utilized to obtain the normalized attention weights $\alpha_t$  corresponding to all hidden states. 

\subsection{Computing probabilities for each of the activities}
Finally, the sum of products of each hidden states  with their attention weights is computed for the final aggregated representation. As a result, important time-stamps are weighed more, and the contributions of less relevant time-stamps become weaker or due to the effects of the attention weights. 
\begin{equation}
v = \sum_{t=1}^{T}\alpha_t h_t.
\end{equation}

Hence, a holistic, contextual vector $v$ of all the time-stamps is provided to the model. After that, as a final step, the context vector $v$ is passed through a softmax layer. This obtains the final results, which are the model-predicted probabilities for each of the activities. 

\subsection{Data Augmentation}

As virtually all Human Activity Recognition (HAR) benchmark datasets are small in size, adding virtual samples to attain data augmentation to increase training data is expected to be able to to improve generalization performance of the model. However, the data augmentation strategies in various machine learning applications, such as small orientation, flipping, cropping, and scaling in computer vision, might not be the most appropriate for multi-sensor channel time-series data. This is because, on one hand, their effectiveness is dataset-dependent and, on the other hand, it is strongly reliant on domain-expert knowledge for appropriate adoption. Even with domain experience, applying these strategies to HAR is non-trivial, as it is difficult to determine to what extent each sensor channel is altered ({\it e.g.}, in re-scaling or flipping) but it still preserves the semantics of the original data label. Here, the {\it mixup} data augmentation  approach~\cite{mu} is a viable option, which generates a virtual sequence by linearly mixing two original sequences with different labels, using different proportions of each sequence. This is because, by generating labels in this way, we can address the label semantics problem posed by handcrafted traditional augmentation approaches. However, due to the linear mixing of two  original sequences with different labels, the generated augmented sequence often appears unnatural and less meaningful, and they can even be fundamentally different from the original sequences. This may cause label semantics alteration and model distortion for test sequences. 

\subsubsection{Temporal CutMix Data Augmentation}\label{cutmix}

\begin{figure}[t] 
  \centering
  \includegraphics[width=\linewidth]{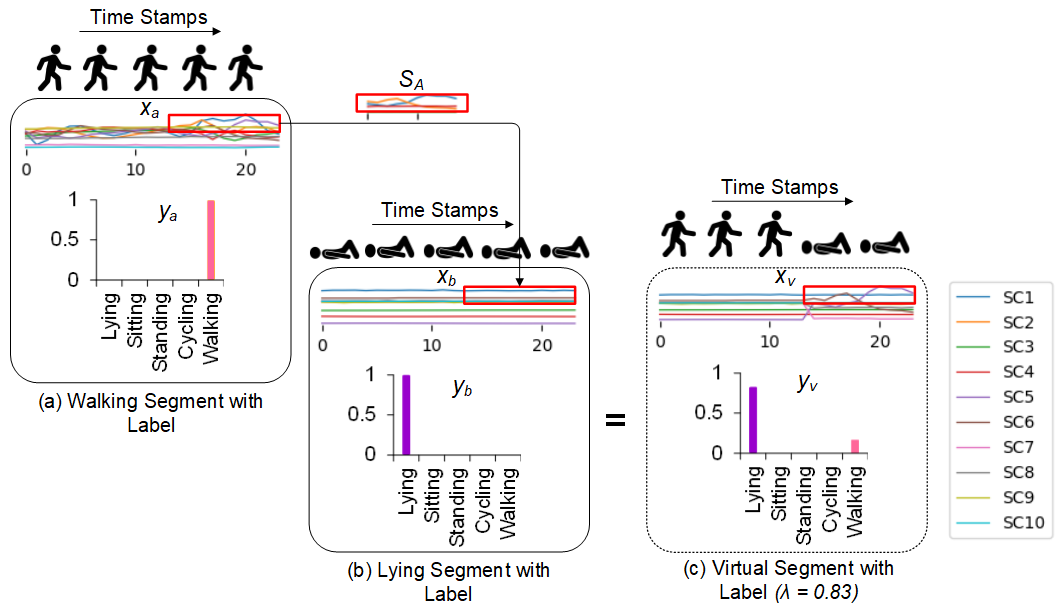}
  \caption{Illustration of our temporal CutMix component generated virtual example: (a) a segment of walking activity ($x_a$) from the training partition with its one-hot encoded label ($y_a$), (b) a training segment of lying activity corresponding to its one-hot encoded label ($y_b$), (c) a newly generated virtual sample with the target label and lambda's ($\lambda \sim BetaDistribution(\alpha, \alpha) $) value indicates the mixing ratio of input segments. The sub-segment $S_A$ from sensors channels $4-8$ with their time-stamps $14-24$ is cropped from $x_a$ and pasted on the corresponding location in $x_b$. Each colour in the sequences $X_a$, $X_b$, and $X_v$, represents a sensor channel, while the range of numbers $0$ to $24$ on the horizontal axis indicates their time-stamps. (Best viewed in color.)}
  \label{fig:cutmix}
\end{figure}
Inspired by the CutMix~\cite{yun2019cutmix} data augmentation method, we first introduce in this Section a \emph{temporal CutMix} data augmentation method for regularizing the datasets to be used in our HAR model.

 We illustrate the temporal CutMix augmentation process through an example in  Figure~\ref{fig:cutmix}, where two differently labeled original training segments $(x_a , x_b)$ are utilized to generate a virtual segment $x_v$  with a label $y_v$. 
 The temporal CutMix algorithm first determines two sub-segments, each from the two original segments $x_a$ and $x_b$. Specifically, we identify the sub-segment $S_A$ with four values $(S_s^C, S_s^T, S_e^C, S_e^T)$. 
 $S_s^C$ and $S_s^T$ indicate the starting sensor channel and starting time-stamp of the sub-segment $S_A$, respectively, while $S_e^C, S_e^T$ represent the ending sensor channel and time-stamp of sub-segment, respectively.  Next, the sub-segment area $S_A$ in $x_a$ is removed, and replaced with the corresponding sub-segment in $x_b$. In the example (see Figure~\ref{fig:cutmix}), the sub-segment $S_A$ (from sensors channels $S_s^C=4$ to $S_e^C=8$ with their time-stamps $S_s^T=14$ to $S_e^T=24$) is cropped from $x_a$ and filled in with the corresponding data from $x_b$). The points of sub-segment $S_A$ are sampled uniformly, {\it i.e.}, these points are chosen arbitrarily from a uniform distribution between the minimum and maximum values, 
\begin{equation}
    \begin{split}
        S_s^C & \sim \text{Uniform Distribution}(0, D), S_e^C =  D\sqrt{1-\lambda} \\
        S_s^T & \sim \text{Uniform Distribution}(0, T), S_e^T =T\sqrt{1-\lambda}.
    \end{split}    
\end{equation}
where $\lambda$, which is derived from the Beta distribution $(\alpha,\alpha)$, indicates the mixing ratio from segment $x_a$ and $x_b$ in the newly generated segment $x_v$.\footnote{Note that $1- \lambda = \frac{S_e^C \times S_e^T}{TD}$.} After generating a new sample $x_v$, we generate its label $y_v$ as,
\begin{equation}
    y_v = \lambda y_a +(1 - \lambda)y_b.
\end{equation}

Temporal CutMix makes it possible to rapidly generate two virtual segments by exchanging the sub-segments in random pairs of training segments from a similar mini-batch at each iteration. 
It 
shows promising results in enhancing the generalization of the HAR model. Moreover, the Temporal CutMix is domain-agnostic and easy to implement. 

\subsubsection{CutMix\textsuperscript{+}} \label{cutmix+}
Despite effectiveness of temporal CutMix in generating new segments from original segments of two different classes, it might generate new segments from two original sequences of intuitively completely different classes, {\it e.g.},  cycling and sleeping.
Consequently, the generated virtual segments appears unnatural and unrecognizable, which can cause ambiguity and hinder the  training and learning performance of the model.
To alleviate this problem, we revise temporal CutMix to CutMix\textsuperscript{+}. Specifically, in each training iteration, we provide the model with one of two original segments, {\it e.g.}, $(x_b, y_b)$ and a newly generated segment $(x_v, y_v )$, but we exclude the segment $(x_a, y_a )$ to reduce the possibility of overfitting. Due to this addition, our loss function is as follows,
\begin{equation}
    \begin{split}
    L_{total} & =L_{CE}^v + L_{CE}^{r_b} = \mathbb{E}[-y_v.\log\overline{y}_v - y_b.\log\overline{y}_b] \\
        & = \mathbb{E}[-[\lambda y_a + (1-\lambda) y_b].\log\overline{y}_v-y_b.\log\overline{y}_b],
    \end{split} \label{10}
\end{equation}
$L_{CE}^v$  defines the evaluation of the CutMix cross-entropy loss of the virtual segment, whereas $L_{CE}^{r_b}$ represents the evaluation of the cross-entropy classification loss of the original segment. $y_a$, $y_b$, and $y_v$ are the actual labels, while $\overline{y}_v$ and $\overline{y}_b$ describe the predicted probability of the virtual segment $x_v$ and the real segment $x_b$, respectively. By utilizing both classification loss and CutMix loss in combination (as shown in Equation~\ref{10}) we improve the performance ability of our model on both virtual segments and original segments.

\section{ALAE-CIE-TAE-CutMix\textsuperscript{+} Framework} \label{sec:EXT}
The proposed ALAE-TAE-CutMix\textsuperscript{+} framework does not directly benefit from the interactions between sensors 
As shown by studies~\cite{sattend2021, ahmad2024hyperhar, ahmad2025virtualhar}, exploiting the interactions between sensors 
enables the HAR model to extract more discriminatory and meaningful representations for an activity. We believe that this finding aligns with the fact that different body parts interact with one another during an activity. These interactions for each activity can be different, and they help differentiate the activities. For example, in the table cleaning activity, unique interactions occur between the right hand and the back of the body. Capturing these kinds of interactions through the attached sensor channels enables the model to extract more distinct representations for activities, ultimately leading to more accurately recognition of activities. Thus, we propose to extend the ALAE-TAE-CutMix\textsuperscript{+} framework with a new end-to-end trainable module, the Cross-Channel Interaction Encoder (CIE), to leverage the interactions between sensor channels, and investigate whether it improves model performance. We refer to this enhanced framework as the new framework ALAE-CIE-TAE-CutMix\textsuperscript{+}. As we have already outlined the modules ALAE, TAE, and CutMix\textsuperscript{+} in detail in Section~\ref{sec:ALAE}, in this section, we present the CIE module and, discuss its design, workflow, and how it is integrated within the ALAE-TAE-CutMix\textsuperscript{+} framework.

\subsection{Cross-Channel Interaction Encoder (CIE)} \label{sec:CIE}

\begin{figure}[] 
  \centering
  \includegraphics[width=\linewidth]{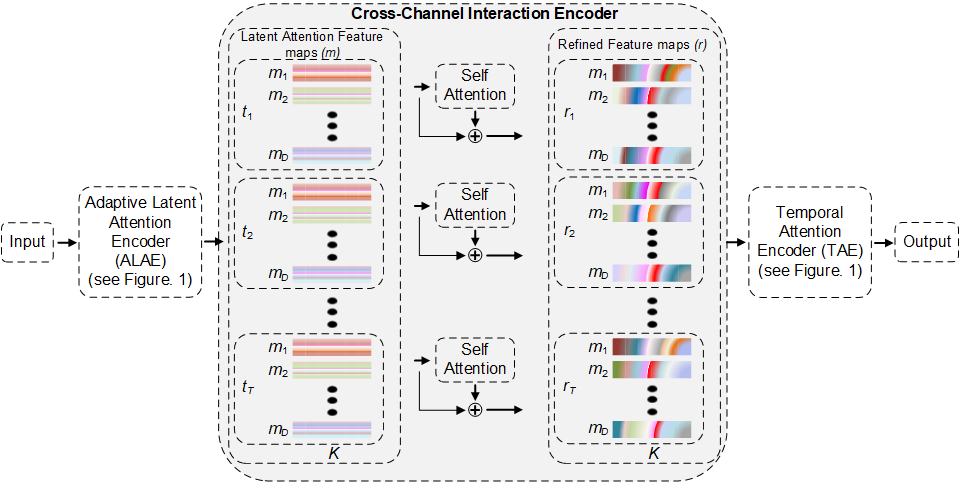}
  \caption{Overview of Cross-Channel Interaction Encoder and showing how it is integrated into ALAE-TAE-CutMix\textsuperscript{+} framework.}
  \label{fig:CIE}
\end{figure}

To learn the interactions between sensor channels, we adopt the CIE module from the study conducted by Abedin~{\it et~al.}~\cite{sattend2021}. The overall architecture of the CIE module is presented in Figure~\ref{fig:CIE}. The module makes use of the self-attention mechanism~\cite{NIPS2017_3f5ee243, SAGAN}, a popular method to learn the relationship between any two variables in the input.
It accepts the sensor channels feature maps $m$ from the ALAE module, learns the interactions between these channels at each time-stamp, and passes them to the TAE module. 
To achieve this, the representation $m(\in \mathbb{R}^{D \times K \times T})$ is passed through the CIE one by one as $t_i (\in \mathbb{R}^{D \times K})$, $t=1, 2, \ldots, T$.  In the first step, we compute the normalized correlations between each dual sensor channel representations $m^d_t$ and $m^{d^{'}}_t$ at every time-stamp $t$,  
\begin{equation}
    a_{t}^{d,d^{'}}=\frac{exp(f(m^d_t)^T g(m^{d^{'}}_t))}{\displaystyle\sum\limits_{d^{'}=1}^D exp(f(m^d_t)^T g(m^{d^{'}}_t))},
\end{equation}
where $a_{t}^{d,d^{'}}$ indicates the normalized correlation value between the features of $d^{th}$ sensor channel and the features of $d^{'th}$ sensor channel at time-stamp $t$. 
After that, the self-attention feature map $o^{d}_t$ at each time-stamp $t$ is produced as the dot product of the normalized correlation values and the actual input sensor channels latent representations $h(m^{d^{'}}_t)$,
\begin{equation}
    o_t^d=\displaystyle\sum\limits_{d^{'}=1}^{D}a_{t}^{d,d^{'}}h(m^{d^{'}}_t).
\end{equation}
Overall, this indicates that the model computes the degrees of interaction between features of the $d^{th}$ sensor channel and the features of each of the $d^{'th}$ sensor channel. Finally, to enable the model to decide whether or not it needs to encode the interaction between a pair of sensor channel with respect to the ongoing activity, a refined representation $r$ is generated. This is achieved through the residual operation~\cite{he2016deep} 
\begin{equation}
    r^d_t = \gamma o^d_t + m^d_t,
\end{equation}
which multiplies the generated attention representation $o^d_t$ with the scalar learning parameter $\gamma$ and then adds it back to the input feature. The parameter $\gamma$  is initialized to 0 and it is gradually increased during the training.  This enables the model to initially focus on learning the simpler interactions between the sensor channels, and then progressively learn the more complex interactions. The purpose of adding the input feature maps $m$ (ALAE output) is to create the connectivity between the ALAE feature maps and self-attention feature maps $o^{d}_t$$-$enabling the model to boost or suppress the ALAE feature maps (sensor channel representations) according to the current ongoing activity. In the aforementioned formulation, the functions $f()$, $g()$, and $h()$ are learnable linear transformation functions. The output of the CIE module produces 
$r = (r_1, r_2, \dots, r_T) \in \mathbb{R}^{D \times K}$. We reshape $r$ into a temporal vector and passes it into the TAE module for learning sequential features.  
\subsection{Overview of complete ALAE-CIE-TAE-CutMix\textsuperscript{+}}
The overall architecture of the complete ALAE-CIE-TAE-CutMix\textsuperscript{+} framework is depicted in Figure~\ref{fig:CIE}. First, we input the CutMix\textsuperscript{+} augmented segment into the ALAE module. The ALAE module  produces the attentive and optimal latent
representation across each sensor channel,  selectively enhancing the representations that are highly related to the ongoing activity. Next, the enhanced sensor channel representations are passed to the CIE module, which learns the interactions across each pair of sensor channels at each time-stamp. Finally, the output of the CIE module is passed to the TAE module, which extracts the global temporal representation of the activity and provides it to the softmax layer to predict the activity. 
\begin{figure}[t]
\centering
  \subfloat[Opportunity]{
        \includegraphics[width=0.50\linewidth]{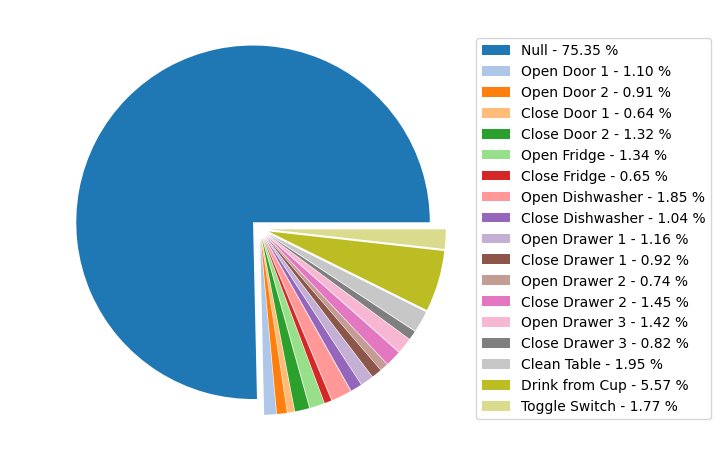}}
  \subfloat[GOTOV]{\includegraphics[width=0.50\linewidth]{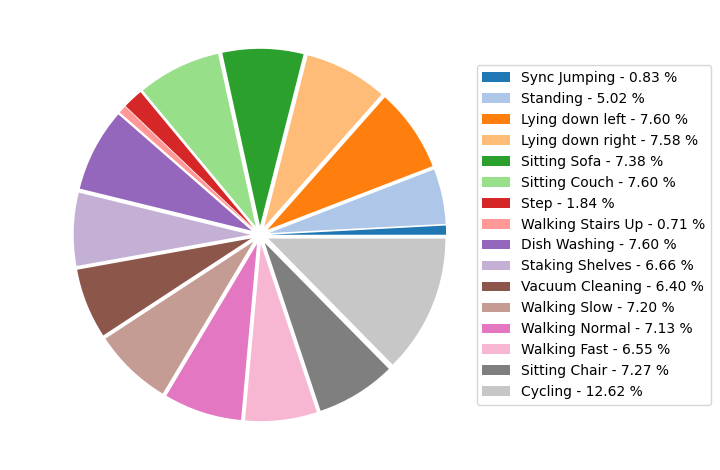}}
  
\subfloat[Skoda]{\includegraphics[width=0.50\linewidth]{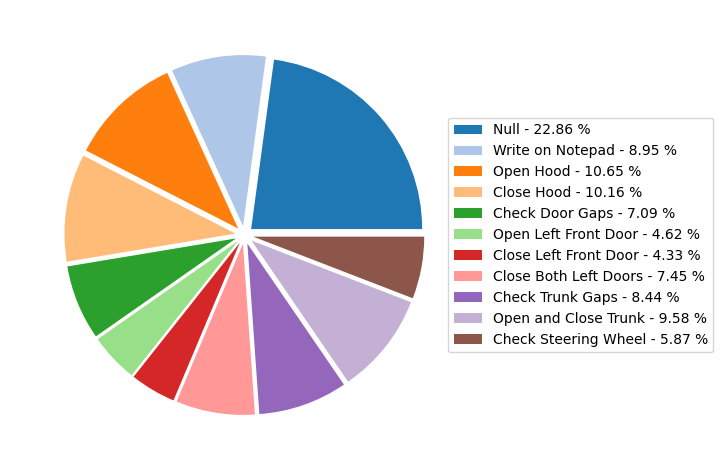}}
 \subfloat[Hospital]{\includegraphics[width=0.50\linewidth]{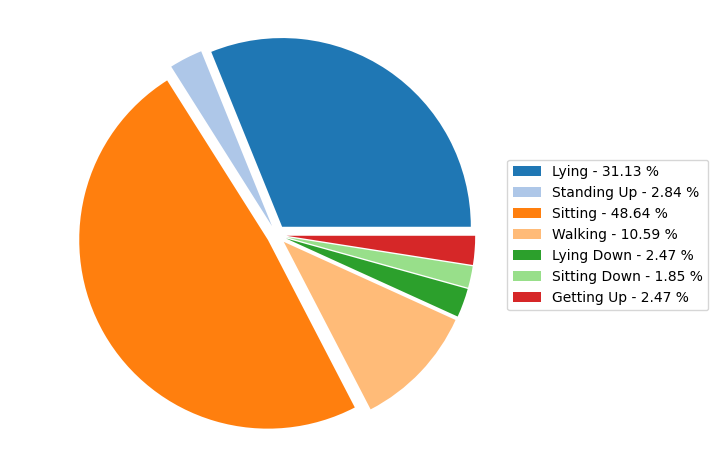}}
\caption{ Pie charts illustrating the four benchmark datasets investigated in our work. The activities listed within these benchmark datasets and their percentage distribution are presented. (Best viewed in color.)}
\label{fig:data}
\end{figure}
\section{Experiments and Results} \label{sec:results}
\subsection{Benchmark Human Activity Recognition Datasets} 
Given that our frameworks focus on addressing the challenges and constraints in recognizing older persons' and patient activities, we use publicly available benchmark datasets, {\it i.e.}, Hospital~\cite{sdense2018}, GOTOV~\cite{gotov} to validate the performance of both proposed models and compare it with state-of-the-art research. The Hospital dataset comprises data on aged patients' activities, and the GOTOV dataset includes data on healthy older adults' activities. These datasets are relevant to elderly care applications. Moreover, to demonstrate the generalizability of the model, we also conduct experiments on highly popular HAR datasets, such as Skoda~\cite{skoda} and Opportunity~\cite{opp}. Skoda contains activities performed by workers on an automotive assembly line, and the Opportunity dataset is a highly unbalanced dataset with a significant bias toward a Null class distribution, accounting for about $75.44\%$ of the data, while most other activities contain data for short gestures. Figure~\ref{fig:data} shows the activities listed in each dataset and their respective distributions. In addition, a statistical overview is provided in Table~\ref{tab:SD}. A brief overview of each dataset is presented below.

\subsubsection{Hospital} The dataset consists of $12$ older, hospitalized patients wearing inertial sensors on their clothing and performing seven different activities. For hold-out evaluation, we follow~\cite{sdense2018}, in which data from the first $8$ and the next $3$ participants are used for training and testing partitions, respectively, while the rest of the data is used for validation.

\subsubsection{GOTOV}
This dataset consists of $9$ sensor channels of data attached to three different body positions, collected from $35$ older adults ($21$ men and $14$ women) over the age of $61$ performing $16$ activities. We exclude six people (four men and two women) from the evaluation due to the unavailability of data from some sensor channels. For hold-out evaluation, the data of three users is used for testing and three for validation, each consisting of two men and one woman, while the data of the remaining users are used for training.
\subsubsection{Skoda} 
This dataset comprises $10$ assembly line activities collected from car manufacturing workers through $60$ sensor channels attached to the right-hand body position. For hold-out evaluation, we follow~\cite{LSTM}, where the first $80\%$ of each label data is used for training, the next $10\%$ for validation, and the remaining for testing.

\subsubsection{Opportunity} 
This dataset is collected through $79$ sensor channels data for $18$ different activities. Four volunteers equipped with wearable sensors carried out the kitchen activities for five different runs. For hold-out evaluation, we replicate~\cite{s2016}, where the data of the $4$th and $5$th runs acquired from subjects $2$ and $3$ are treated as a test, and the data of the $2$nd run collected from subject one is assigned to the validation section, and the remaining data is considered training data.
\begin{table}[t]
  \caption{Statistical overview of the datasets.}
  \label{tab:SD}
  \begin{center}
  \begin{tabular}{lcccccl}
    \hline
    Dataset & \# Subjects & \# Sensor Channels & \# Activities & \# Training Segments\\
    \hline
    Hospital & 12 & 6 & 7 & 4607 \\
   
    GOTOV & 29 & 9 & 16 & 291723 \\
    
    Skoda & 1 & 60 & 10 & 15548 \\
    
    Opportunity & 4 & 79 & 18 & 54246 \\
    \hline
\end{tabular}
\end{center}
\end{table}

\subsection{Evaluation} \label{EP}
For a fair comparison, we use the same evaluation approach ({\it e.g.}, hold-out), performance metrics, and standard training and testing partitions used in our state-of-the-art studies~\cite{s2016, LSTM,sattend2021, satt2018}.  Following~\cite{LSTM, sattend2021, satt2018}, training data is divided into segments using a sliding window technique before feeding the data into the learning model.  During training, we utilize $24$ time-stamps in each segment with $50\%$ overlap between consecutive windows, for example, a new window $j$ overlaps with $12$ samples of window $j-1$. While in testing, following ~\cite{sattend2021, satt2018}, a more practical and realistic technique, sample-by-sample evaluation is employed in which prediction is made for each test sample. As our utilized HAR datasets are imbalanced (as shown in Figure~\ref{fig:data}), thus following studies ~\cite{s2016, LSTM,sattend2021, satt2018}, we also use the F-score
\begin{equation}
    \mbox{F-score}=2 \times \frac{precision \times recall}{precision + recall},
\end{equation}
 evaluation metric to report our models performance. In addition, confusion matrices are reported for the proposed frameworks to show the activity-wise model’s performance. 
\subsection{Experimental Setup and Hyperparameter Settings} \label{sec:HS}
We implement our proposed frameworks using TensorFlow Keras and conduct all the experiments on an Nvidia RTX $3090$ GPU. Dropout is employed on the ALAE and TAE modules to avoid overfitting. Specifically, we choose $0.5$ as the dropout probability on the attention module $M's$ first and second dense layers, while in the TAE module, $0.5$ dropout and $0.5$ recurrent dropouts are used on the LSTM first layer, and $0.5$ and $0.9$ as the LSTM second layer, respectively. Our model learning parameters are optimized for up to $300$ iterations in an end-to-end training manner using backpropagation and the cross-entropy loss function on a mini-batch size of $256$. Adam is used as a learning parameter optimizer, with a learning rate of $0.001$ that decreases by $0.9$ every ten epochs. For CutMix\textsuperscript{+} data augmentation we utilize $\alpha=0.3$. All the hyperparameters are kept the same across all four utilized datasets.

\subsection{Comparison with the State-of-the-Art HAR studies} \label{sec:c-SOTA}
 In this section, we compare the performance of the proposed frameworks with state-of-the-art HAR studies. We conduct a pool of experiments, each of which involves the random initialization of the learning parameters. Initially, we follow the traditional way of comparing the performance, in which we report the best experiment result among all conducted experiments to show the effectiveness of the proposed frameworks. We observe from the pool of experiments that the simpler framework's best experiment result is relatively better than the CIE-enhanced framework's best result. However, the results of most of the other experiments of the CIE-enhanced framework are relatively better and more stable (the results range was quite small). Given this situation, we go a step further and examine the reliability and robustness of the proposed frameworks by presenting a performance comparison based on average results and the standard deviations of all the experiments.  

 \subsubsection{Effectiveness}
To show the effectiveness of our proposed frameworks, we compare our results with five leading-edge HAR studies that proposed one of the most widely recognized
models for HAR (listed in Table~\ref{tab:SOTAcomp}). For the Hospital, Skoda, and Opportunity datasets, the results of baseline studies~\cite{sCNNLSTM2016, s2016} are directly quoted from~\cite{sattend2021}, whereas the results of other baseline studies are obtained from~\cite{LSTM, sattend2021, satt2018}. As no results are available for the GOTOV dataset, we utilize the published code of the baseline studies to obtain the performance data.
\begin{table}[]
\caption{An $\text{F-score}$ based performance comparison between the proposed frameworks  and the baseline studies.}
\begin{center}
\begin{tabular}{p{0.000005cm}cccccc}
\hline
& HAR Study & Hospital & GOTOV & Skoda & Opportunity \\
\hline
 \multirow{5}{*}{\rotatebox[origin=c]{90}{Baselines}} & LSTM Learner Baseline~\cite{LSTM} & 62.7 & 61.1 & 90.4 & 65.9\\
 & DeepConvLSTM~\cite{sCNNLSTM2016} &	62.8 & 66.9 & 91.2 &	67.2 \\
  & b-LSTM-S~\cite{s2016} &	63.6 & 63.9 & 92.1 &	68.4\\
  & Att. Model~\cite{satt2018} & 64.1 & 70.7 & 91.3 &	70.7\\
   & {Attend and Discriminate}~\cite{sattend2021} & 66.6 & 76.2 &	92.8 &	74.6\\
 \hline
 &\textbf{ALAE-TAE-CutMix\textsuperscript{+}} & \textbf{70.5} & \textbf{79.4} &	\textbf{94.8} &	\textbf{75.1} \\
 &\textbf{ALAE-CIE-TAE-CutMix\textsuperscript{+}} & \textbf{70.3} & \textbf{79.7} &	\textbf{94.8} &	\textbf{73.4} \\
 &\textbf{Improvement of ALAE-TAE-CutMix\textsuperscript{+} over~\cite{sattend2021}} & \textbf{(5.86\%)} & \textbf{(4.12\%)} &	\textbf{(2.16\%)} &	\textbf{(0.68\%)} \\
\end{tabular}
\label{tab:SOTAcomp}
\end{center}
\end{table}

\begin{figure}[!ht]
\centering
\begin{subfloat}
  \centering
  \includegraphics[width=0.49\linewidth]{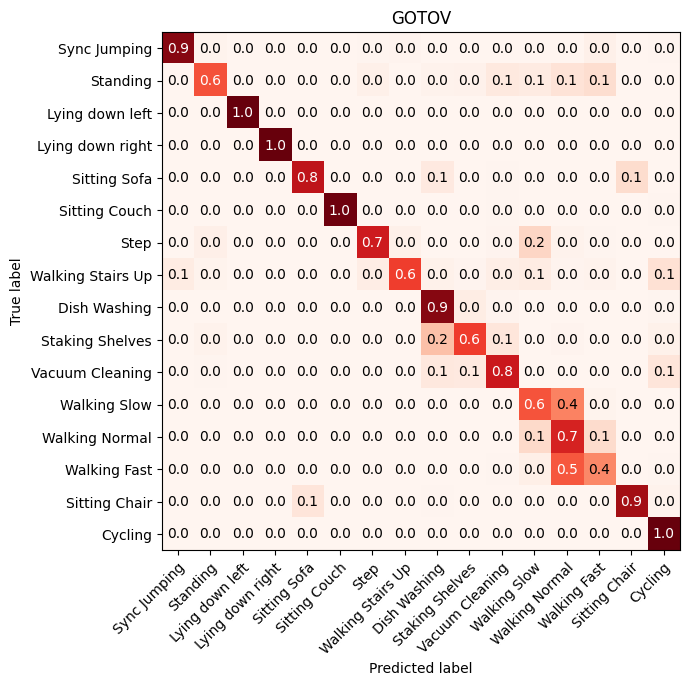}
  \label{fig:sub1}
\end{subfloat}%
\begin{subfloat}
  \centering
  \includegraphics[width=0.49\linewidth]{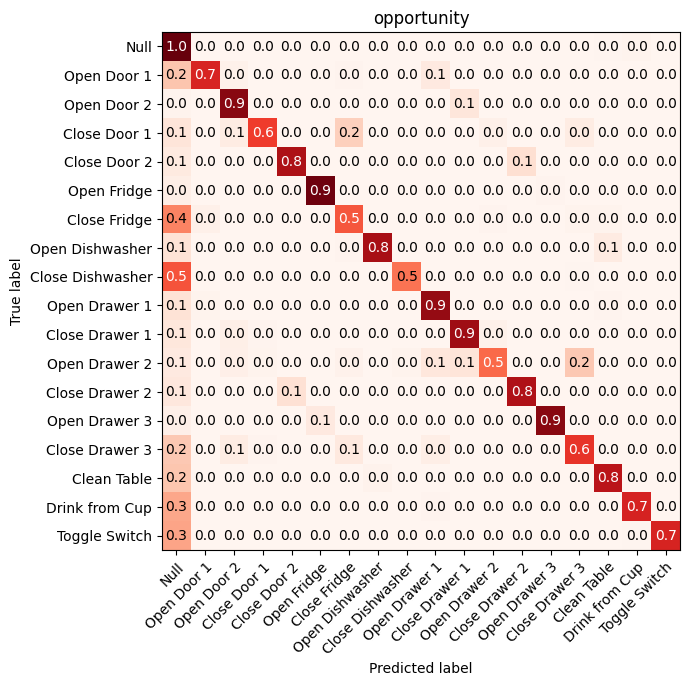}
  \label{fig:sub2}
\end{subfloat} 
\begin{subfloat}
  \centering
  \includegraphics[width=0.49\linewidth]{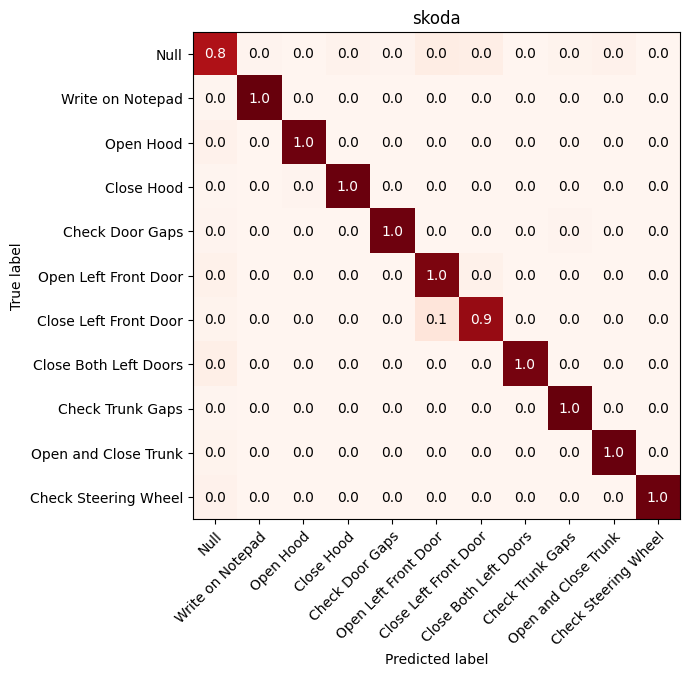}
  \label{fig:sub2}
\end{subfloat}
\begin{subfloat}
  \centering
  \includegraphics[width=0.49\linewidth]{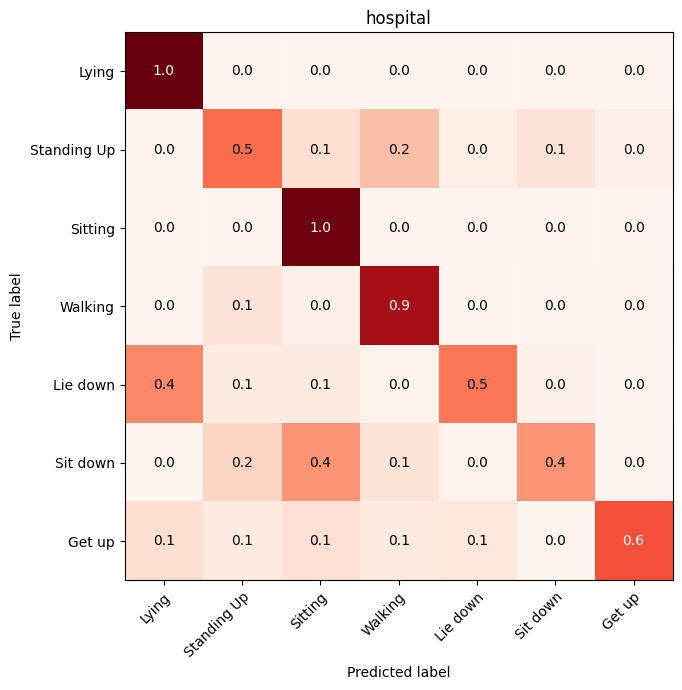}
  \label{fig:sub2}
\end{subfloat}
\caption{Confusion matrices to show class-wise recognition performance  of ALAE-TAE-CutMix\textsuperscript{+} framework achieved on the hold-out test splits across the four benchmark datasets.}
\label{fig:con}
\end{figure}

\begin{figure}[!ht]
\centering
\begin{subfloat}
  \centering
  \includegraphics[width=0.49\linewidth]{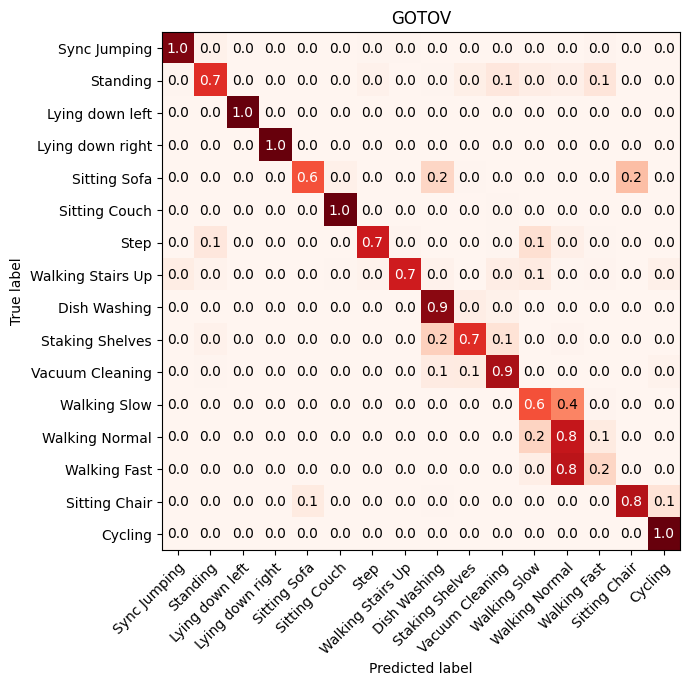}
  \label{fig:sub1}
\end{subfloat}%
\begin{subfloat}
  \centering
  \includegraphics[width=0.49\linewidth]{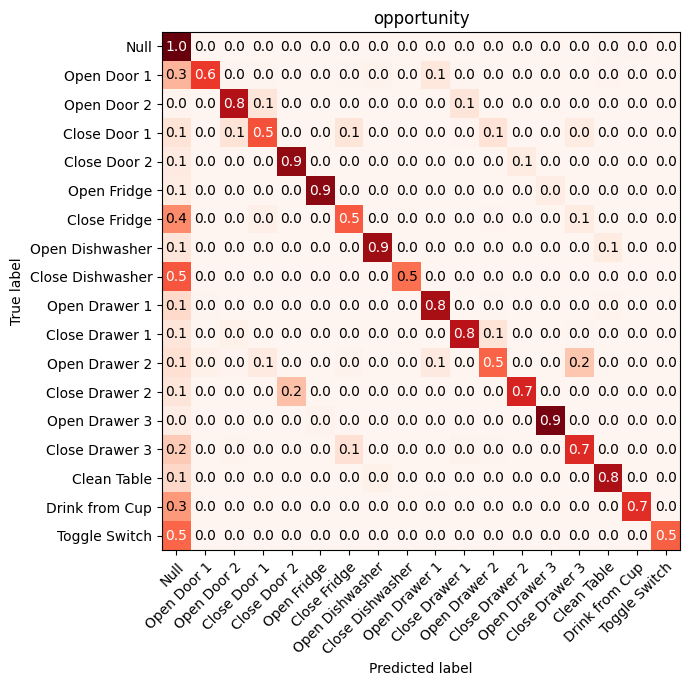}
  \label{fig:sub2}
\end{subfloat} 
\begin{subfloat}
  \centering
  \includegraphics[width=0.49\linewidth]{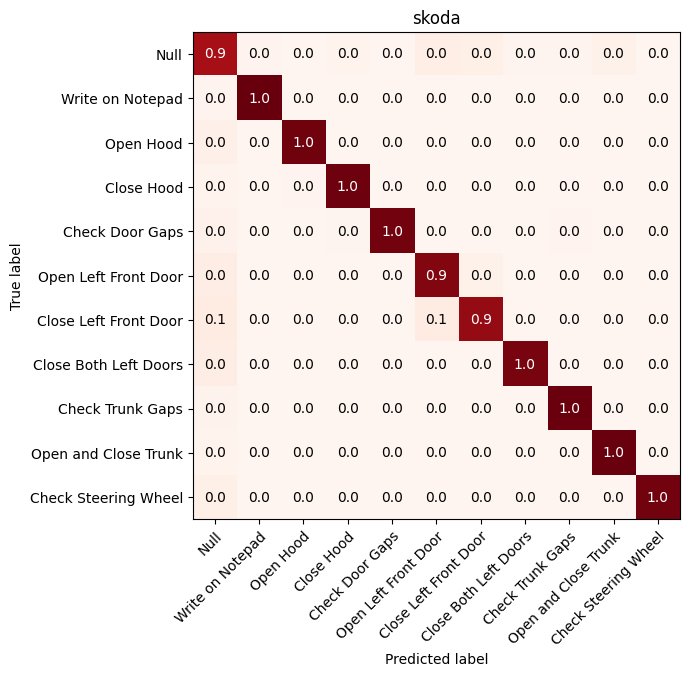}
  \label{fig:sub2}
\end{subfloat}
\begin{subfloat}
  \centering
  \includegraphics[width=0.49\linewidth]{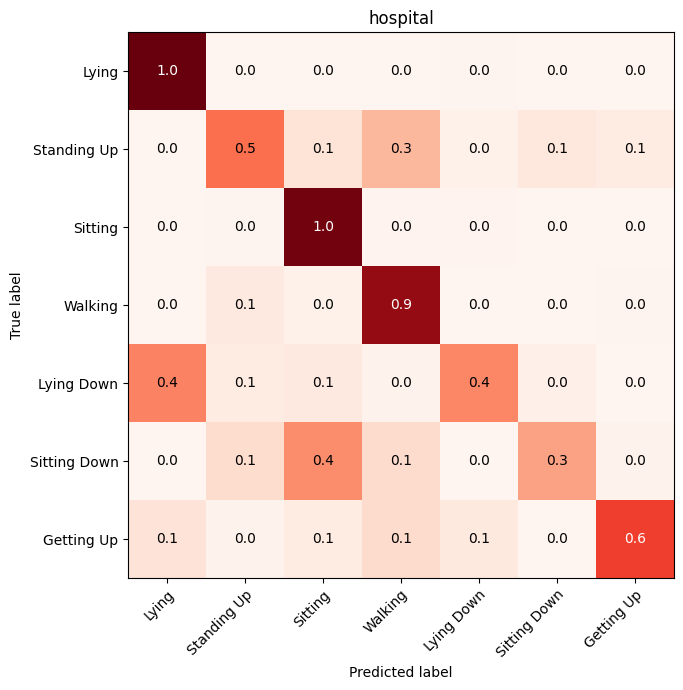}
  \label{fig:sub2}
\end{subfloat}
\caption{Confusion matrices to show class-wise recognition performance of ALAE-CIE-TAE-CutMix\textsuperscript{+} framework achieved on the hold-out test splits across the four benchmark datasets.}
\label{fig:E_con}
\end{figure}

 From the data in Table~\ref{tab:SOTAcomp}, it is clear that both proposed models, with and without CIE, outperform all baseline studies across all datasets. However, in this scenario of result reporting, our simpler model (without CIE) is generally more effective $-$ surpassing the best-performing state-of-the-art~\cite{sattend2021} by $5.86\%$, $4.12\%$, $2.16\%$, and $0.68\%$ on the Hospital, GOTOV, Skoda and Opportunity datasets, respectively. Even though the Hospital and GOTOV datasets contain inter-class similarity problems and the Hospital dataset is small and imbalanced (unequal distribution of classes in the training dataset, see Figure~\ref{fig:data}), it should be noted that the proposed models achieve a significant performance improvement in recognizing the activities of older people (both hospitalized and non-hospitalized). Furthermore, in the case of the Skoda dataset, when recognizing the activity performance of assembly line workers, there is a considerable improvement over the state-of-the-art ~\cite{sattend2021}. Notably, for the Opportunity dataset, our ALAE-TAE-CutMix\textsuperscript{+} model  also outperforms state-of-the-art despite a high level of class imbalance and greater diversity of activities. For further insights and a better understanding of the performance of our model, we provide the class-wise recognition performance through confusion matrices across all datasets in Figures~\ref{fig:con} and~\ref{fig:E_con} for the simpler framework and CIE-enhanced framework, respectively. For the Opportunity dataset, most activities are confused with ``Null'' class activities. This was expected, as the Null class holds an infinite number of unknown activities information---more than $75\%$ of the total data (see Figure~\ref{fig:data}(a)). This is one of the research problems in the field of HAR. 
\begin{figure}[h]
\centering
  \subfloat[]{\includegraphics[width=\textwidth, height=5.7cm,keepaspectratio]{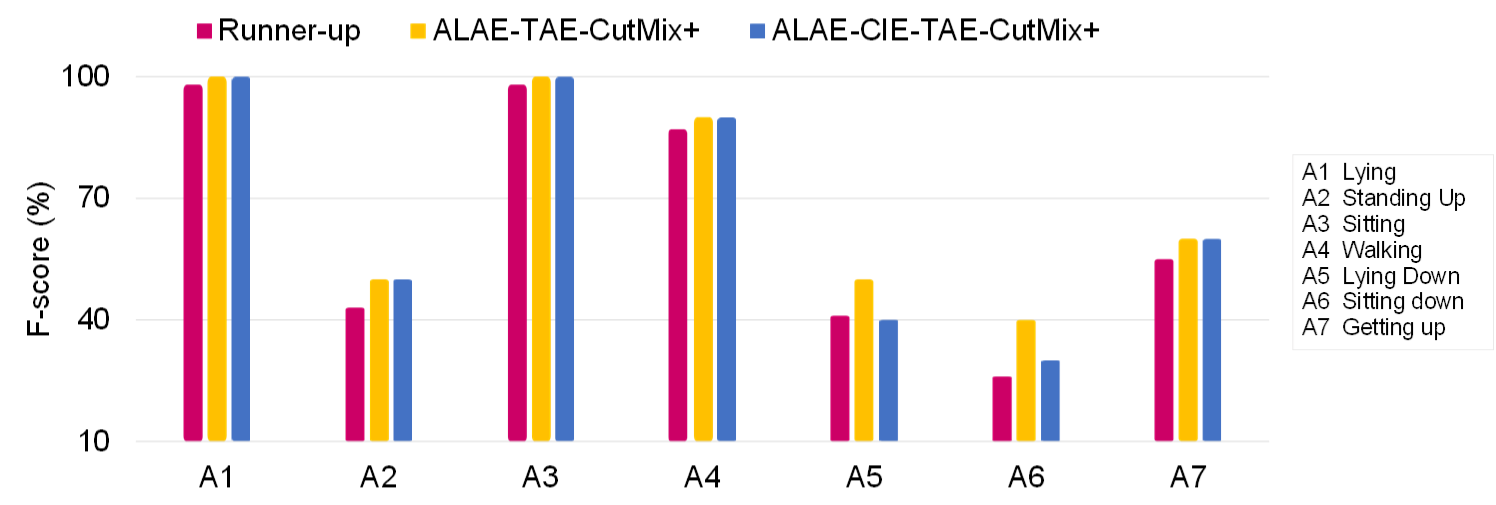}}\quad
  \subfloat[]{\includegraphics[width=\textwidth]{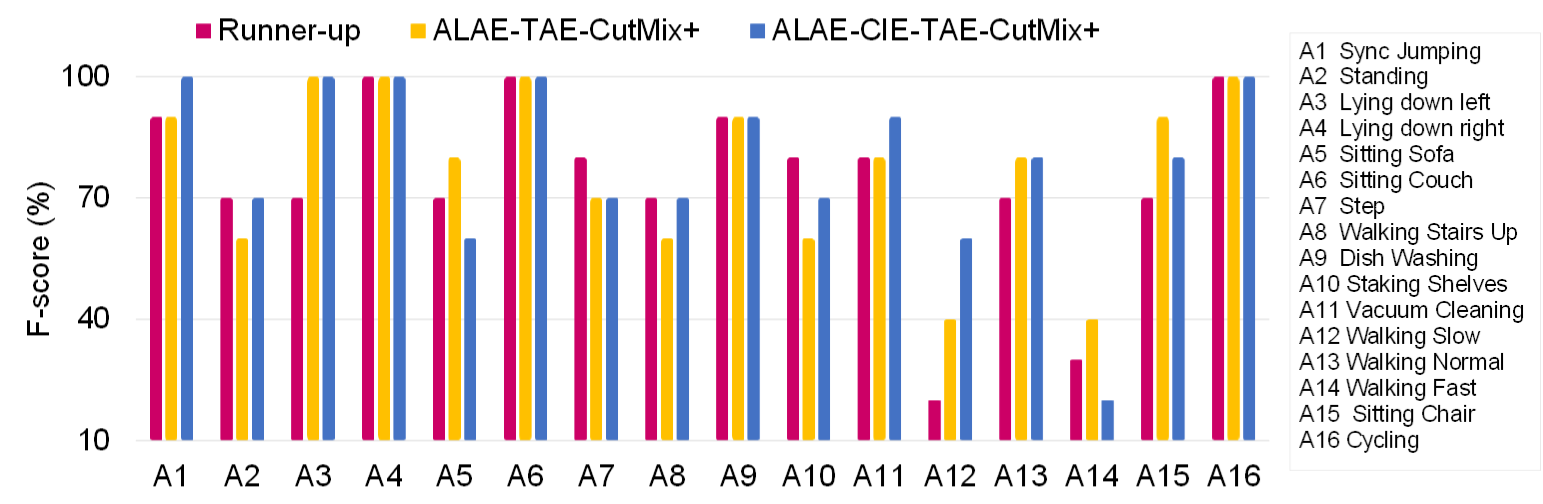}}
\caption{Activity-wise recognition performance comparison among both proposed models and the runner-up Attend and Discriminate model~\cite{sattend2021} on (a) Hospital and (b) GOTOV dataset.}
\label{fig:graph-com}
\end{figure}
To show the activity-wise improvements of our models, we conduct the  performance comparison between our models and the runner-up model~\cite{sattend2021} across Hospital and GOTOV datasets in Figure~\ref{fig:graph-com}. The proposed models achieve superior performance. These datasets include inter-class similar activities' complexities---lacking discriminatory information across activities. In this scenario, developing a model that obtains desirable performance is not a trivial task.
Hence, substantial improvement of our methods  on the aforementioned datasets is noteworthy, particularly for inter-class similar activities ({\it e.g.}, see sitting-down and lying-down activities in the Hospital dataset, while the GOTOV dataset includes sitting on a sofa and sitting on a chair, and three types of walking (slow, normal, and fast)). This is strong evidence of enhancing discriminatory information, which was our primary motivation for developing the ALAE module.

\subsubsection{Reliability and Robustness} 
To obtain the complete picture of the performance of our proposed frameworks, we also show how reliable and robust they are. For reliability, we see how accurately the models perform in each experiment. In the robustness, we see the consistency in the performance of models in different experiments. To measure the model's reliability, we report the average result of all the conducted experiments. Conversely,  to measure the model’s robustness, we report the standard deviations of all the conducted experiments. In Table~\ref{tab:aver-SOTAcom}, we report the results of all the experiments using the CIE-enhanced ALAE-CIE-TAE-CutMix\textsuperscript{+} framework, that of the simpler ALAE-TAE-CutMix\textsuperscript{+} framework without the CIE module, and that of the current state-of-the-art model (Attend and Discriminate~\cite{sattend2021}). \footnote{Since the results of the Attend and Discriminate study are unavailable, we use publicly available code to obtain its results.} 

Based on the results, it is evident that the computed average scores of the CIE-enhanced framework are significantly higher, while the obtained standard deviations are generally lower than those of the proposed simpler framework and current state-of-the-art  Attend and Discriminate framework. Hence, this shows that the CIE-enhanced framework is more reliable and robust than other frameworks. In addition, compared to the Attend and Discriminate model, our simpler model (without the CIE module) also offers more robustness and reliability.

\subsection{Ablation Study}
Given that we integrate several new modules (CutMix\textsuperscript{+}, ALAE, CIE and TAE) into the proposed HAR framework, it is therefore mandatory to see the effects and contributions of each component in the proposed framework. To do this, we conduct an ablation study on the Hospital dataset, and report the contribution of each module alone and in conjunction with other proposed modules in Table~\ref{tab:AB}. 
\begin{table}[]
\caption{ Performance comparison between the proposed ALAE-CIE-TAE-CutMix\textsuperscript{+}, ALAE-TAE-CutMix\textsuperscript{+} and the current best performing model based on average results and the standard deviations of all experiments.}
\begin{center}
\begin{tabular}{ccccccc}
\hline
& HAR Study & Hospital & GOTOV & Skoda & Opportunity \\
\hline
 & {Attend and Discriminate}~\cite{sattend2021} & 64.8 $\pm$ 0.8  & 75.3 $\pm$ 0.9 &  92.2 $\pm$ 0.2	 & 70.6 $\pm$ 2.8	\\
 &ALAE-TAE-CutMix\textsuperscript{+} & 67.7 $\pm$ 1.1 & 77.8 $\pm$ 0.7 &  94.5 $\pm$ 0.1 &	71.3 $\pm$ 0.4\\ 
 &ALAE-CIE-TAE-CutMix\textsuperscript{+} & 68.7 $\pm$ 0.7  & 78.4 $\pm$ 0.2 & 94.6 $\pm$ 0.1 & 71.7 $\pm$ 1.0	 \\
\hline
\end{tabular}
\label{tab:aver-SOTAcom}
\end{center}
\end{table}

 We start by removing all other modules from the proposed framework, leaving only the TAE module's LSTM layers and the classification layer. As a result, it becomes the similar model ({\it i.e.}, Baseline LSTM Learner) proposed in the study~\cite{LSTM}, which we use as our reference model for the ablation study. We add each module with the reference model in their respective places (as chosen in the proposed framework) alone and in different combinations to see how much the addition of the modules improves the performance compared to the reference model.
Unsurprisingly, eliminating all new components except the reference model makes the performance largely drop to $62.7\%$ (as expected, the LSTM based learning model~\cite{LSTM} also reported $62.7\%$ (see Table~\ref{tab:SOTAcomp})). In contrast, when all new modules (CutMix\textsuperscript{+}, ALAE, and TAE) are utilized together, the F-score performance improves significantly to $70.5\%$. However, it is interesting when CIE is used with all the new modules, the performance is slightly dropped, which we discuss shortly.

The addition of only the CutMix data augmentation with the reference model obtains an impressive $3.5\%$ performance gain over the baseline (from $62.7\%$ to $66.2\%$), showing its effectiveness in generalizing the HAR model. Moreover, this implies that CutMix's generated multi-channel samples successfully augment training data and enhance the generalization of the learned activity's key representations to test sequences. Similarly, integrating the ALAE module alone achieves a $3.5\%$ improvement over the LSTM reference model (from $62.7\%$ to $66.2\%$). This shows the ALAE module contribution toward enhancing discriminatory information for each activity. 

As expected, the integration of ALAE with CutMix (CutMix + ALAE) significantly boosts performance by $7\%$ over the baseline (from $62.7\%$ to $69.7\%$). Further, the inclusion of TAE with other modules marginally improves model performance. Additionally, we see more performance improvements and faster model convergence when integrating CutMix\textsuperscript{+} with other modules to train the model on both virtual and raw sequences.

Adding the CIE module with all the proposed modules has a slightly negative impact on the model performance. However, it ought to be noted that the CIE module has a significant impact in terms of increasing the robustness and reliability of the model as illustrated in Section~\ref{sec:c-SOTA}.

We also conduct an ablation study on the GOTOV dataset, as its characteristics differ significantly from those of the Hospital dataset. For example, GOTOV data comes from healthy aged subjects, whereas the Hospital dataset data is from older patients; GOTOV includes data from a larger number of sensor channels, and the GOTOV dataset is around $63.30$ times more data than the Hospital dataset.  The results are shown in Table~\ref{tab:AB_gotov}. It is important to note that the addition of the ALAE module is consistently improving the performance of the model. As shown in the Table~\ref{tab:AB_gotov}, whenever we use the ALAE module (alone and also in combination with other modules), it is significantly contributes to the performance enhancement of the model. This shows the substantial effectiveness of the ALAE module toward extracting the discriminatory features across each activity.
\begin{table}[t]
  \caption{The results of ablation study on the Hospital dataset.}
  \label{tab:AB}
  \begin{center}
  \begin{tabular}{lccl}
    \hline
    HAR Model & F-score\\
    \hline
    LSTM Learner Baseline & 62.7\\
    Ours (CutMix) & 66.2 \\
    Ours (ALAE) & 66.2 \\
    Ours (CutMix\textsuperscript{+}) & 66.7 \\
    Ours (CutMix + TAE) & 66.7\\
    Ours (ALAE + TAE) & 66.8 \\
    Ours (CutMix\textsuperscript{+} + TAE) & 66.9\\
    Ours (CutMix + ALAE) & 69.7 \\
    Ours (CutMix + ALAE + TAE) &  69.9 \\
    \hline
    \textbf{Ours (CutMix\textsuperscript{+} + ALAE + TAE)} & \textbf{70.5} \\
    \textbf{Ours (CutMix + CIE + ALAE + TAE)} & \textbf{70.3} 
\end{tabular}
\end{center}
\end{table}

\begin{table}[t]
  \caption{The results of ablation study on the GOTOV dataset.}
  \label{tab:AB_gotov}
  \begin{center}
  \begin{tabular}{lccl}
    \hline
    HAR Model & F-score\\
    \hline
    LSTM Learner Baseline & 61.1\\
    Ours (CutMix) & 64.8 \\
    Ours (ALAE) &  73.7 \\
    Ours (CutMix\textsuperscript{+}) & 63.3  \\
    Ours (CutMix + TAE) & 64.9 \\
    Ours (ALAE + TAE) & 74.4 \\
    Ours (CutMix\textsuperscript{+} + TAE) & 65.0\\
    Ours (CutMix + ALAE) & 78.9 \\
    Ours (CutMix + ALAE + TAE) & 78.4  \\
    \hline
    \textbf{Ours (CutMix\textsuperscript{+} + ALAE + TAE)} & \textbf{79.4} \\
    \textbf{Ours (CutMix + CIE + ALAE + TAE)} & \textbf{79.7}  
\end{tabular}
\end{center}
\end{table}
We also investigate how hand-crafted traditional data augmentation methods affect the performance of the learning models (ALAE-TAE and ALAE-CIE-TAE). We adopt the recently explored methods that include scaling, jittering, and magnitude warping from the study conducted by Um~et~al.~\cite{41} and apply them on the proposed models. The results  are detailed in Table~\ref{tab:abl-DA}. Scaling changes the segment data magnitude by  multiplying arbitrarily scalar number with the sensor channel data samples. Jittering introduces simulated noise to the sensor channel data. Finally, magnitude warping introduces subtle variations in the magnitude of each data sample by convolving the data segment with a smooth curve whose value fluctuates around one.

Based on the findings, we find that CutMix and its variants exhibit a significant performance advantage over all traditional handcrafted data augmentation methods. From Table~\ref{tab:abl-DA}, it can be observed that in most cases, the traditional augmentation methods highly degrade the performance of the models, particularly on the GOTOV dataset. We believe that this may be due to random alteration of each sensor channel data without considering its label's knowledge.  Consequently, the artificially generated sequences suffer from the label alteration problem ({\it i.e.}, the data is altered in a way that becomes semantically different from the original activity label) and training the model on this artificially generated sequence misleads the model, which ultimately causes the model distortion for test sequences. For example, consider scaling data augmentation: When the sequence for the slow walk activity is augmented by scaling with a larger generated number, the data magnitude for each sensor channel increases. Consequently, the data pattern diverges semantically and may look like a pattern of fast walking or jogging activity. Moreover, when the model is trained on this augmented sequence with the original label (slow walk), it can lead to the model being misleading. This problem can be reduced if these data augmentations are carefully tuned and applied~\cite{41}. However, carefully tuning is not a straightforward task due to the multiple challenges involved, for example, what and how many activities are to be recognized, the similarity of inter-class activities, the variation in intra-class activities, and the number of sensor channels and their characteristics {\it etc.}~\cite{sattend2021}. 
\begin{table}[t]
\centering
\caption{The F-score based comparison between proposed CutMix and its variants with different data augmentation methods on Hospital and GOTOV dataset.}
\begin{tabular}{ccccc}
\hline
Data Augmentation          & \multicolumn{2}{c}{Hospital} & \multicolumn{2}{c}{GOTOV} \\

                            & ALAE-TAE    & ALAE-CIE-TAE   & ALAE-TAE  & ALAE-CIE-TAE  \\ 
\hline
No Augmentation            & 68.2        & 66.7           & 72.4      & 70.6         \\
Scaling                     & 67.1        & 66.8           & 72.7      & 75.5          \\
Jittering                   & 67          & 67.2           & 71.3      & 75.7          \\
Magnitude Warping           & 68.5        & 67.9           & 71.0        & 74.7          \\
\hline
\textbf{CutMix with No Label Mixing} & \textbf{68.2 }       & \textbf{68.3}           & \textbf{76.1 }     & \textbf{78.3}          \\
\textbf{CutMix}                      & \textbf{69.9}        & \textbf{70.2}           & \textbf{78.8}      & \textbf{78.9}          \\\textbf{}
\textbf{CutMix\textsuperscript{+}    }                & \textbf{70.5 }       & \textbf{70.3}           & \textbf{79.6 }     & \textbf{79.7}    \\     
\hline
\end{tabular}
\label{tab:abl-DA}
\end{table}

Importantly, compared to the handcrafted data augmentation methods, CutMix generates novel augment segments by considering the original data (not any noise or random values) and the labels. In this manner, it largely alleviates the problem of changing the label semantics, leading to more accurate augmentation for HAR. This can be validated by comparing the results over no augmentation in Table~\ref{tab:abl-DA}: CutMix significantly boosts, while handcrafted methods either reduce or minimally improve the model performance.

In addition, to further strengthen our observation that handcrafted augmentation methods can alter the semantic representation of the original activity label in generated augmented samples, we conduct additional experiments with the CutMix data augmentation. In these experiments, we only mix the two input activities (according to the mixing ratio) to generate the augment segments without mixing their corresponding labels. The majority ratio activity input label is assigned to the newly generated augmented segment. According to the findings presented in Table~\ref{tab:abl-DA}, by comparing the results of the CutMix and CutMix\textsuperscript{+}, we observe that the model's performance is negatively impacted when the labels from both activities are not mixed according to the mixing ratio. This effect is observed even though we ensure that the likelihood of label transformation semantics from the original activity in the newly generated segment is minimal, which is achieved by transforming the original segment with a small amount of data from the other activity and retaining the original label to the newly transformed augmented segment. Nevertheless, CutMix with no label mixing performs better than no augmentation and handcrafted methods in most cases. This improvement could be due to the mixing of minor parts from the second activity segment, potentially contributing to the model's regularization.    

\subsection{Impact of Less Amount of Training Data on Model Performance}
\begin{figure}[t]
\centering
\subfloat[Hospital dataset]{
  \includegraphics[width = 3.6cm] {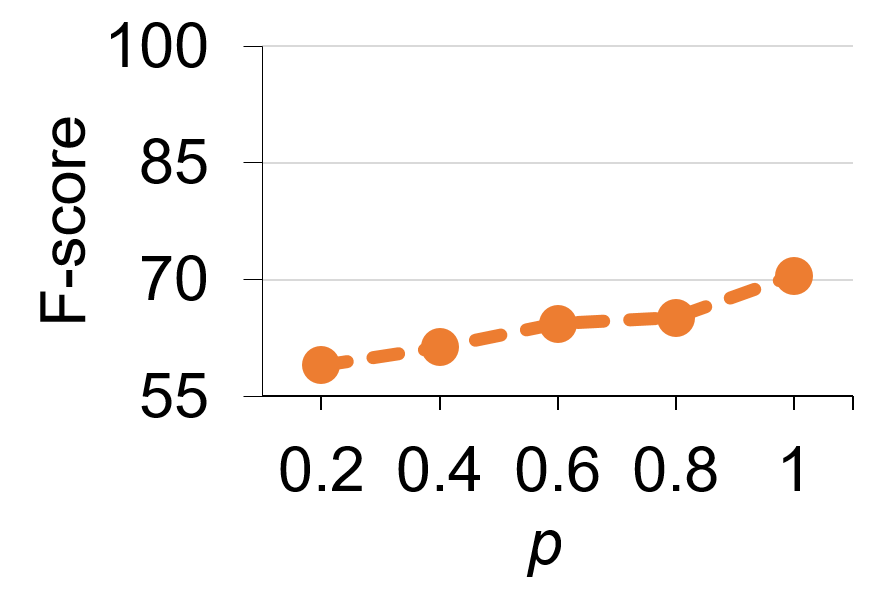}
  }
  \subfloat[GOTOV dataset]{
  \includegraphics[width = 3.6cm] {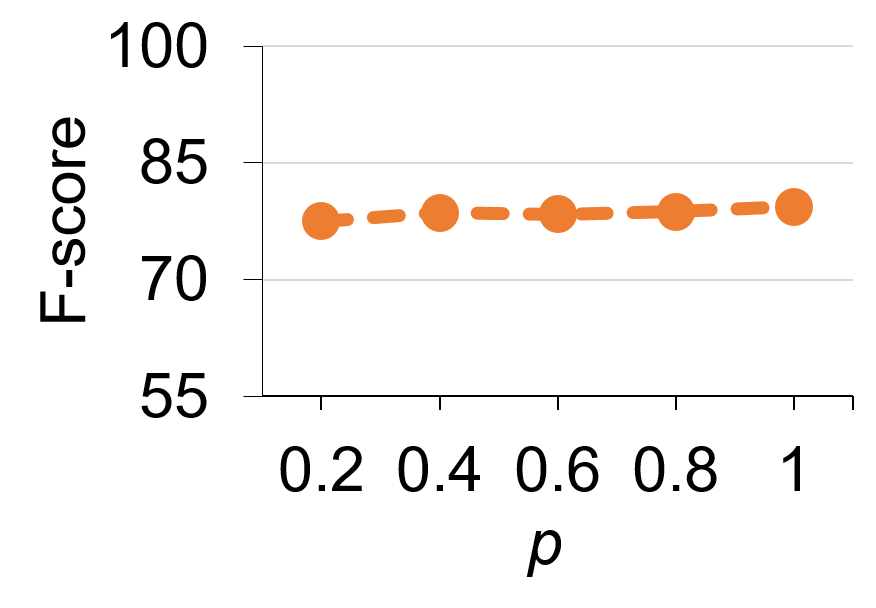}
  }
  \subfloat[Skoda dataset]{
  \includegraphics[width = 3.6cm] {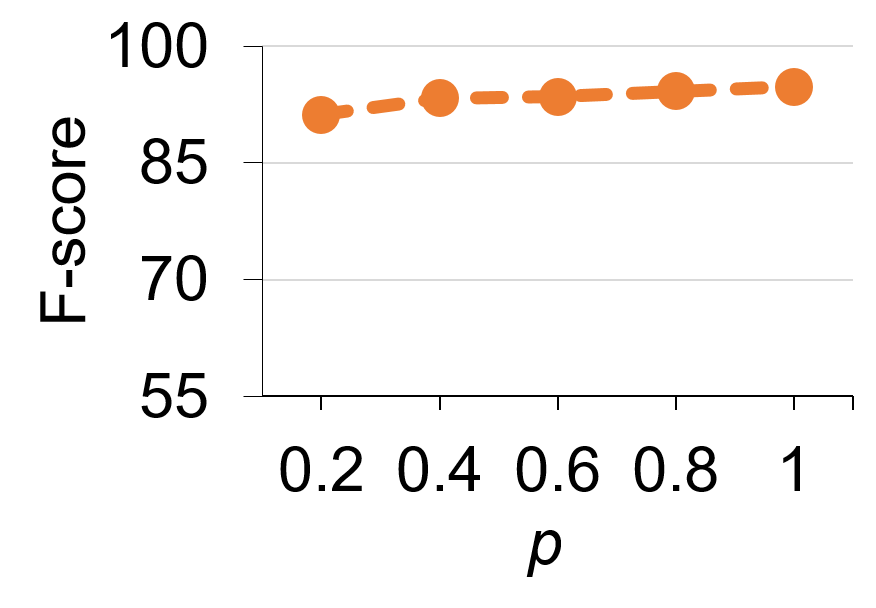}
  }
  \subfloat[Opportunity dataset]{
  \includegraphics[width = 3.6cm] {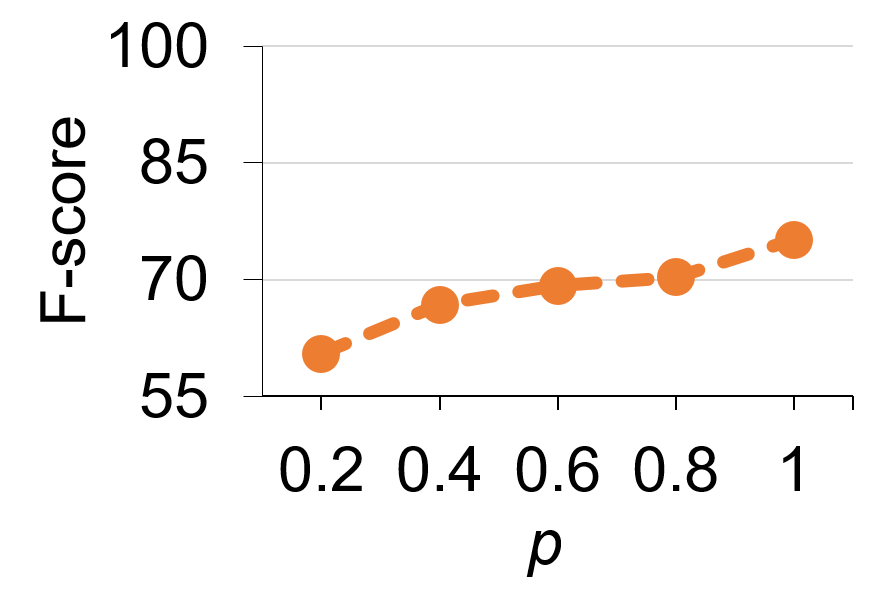}
  }
\caption{
The performance analysis of the proposed simpler framework on limited training data across different datasets. \label{fig:lim-ALAE}}

\end{figure}

\begin{figure}[t]
\centering
\subfloat[Hospital dataset]{
  \includegraphics[width = 3.6cm] {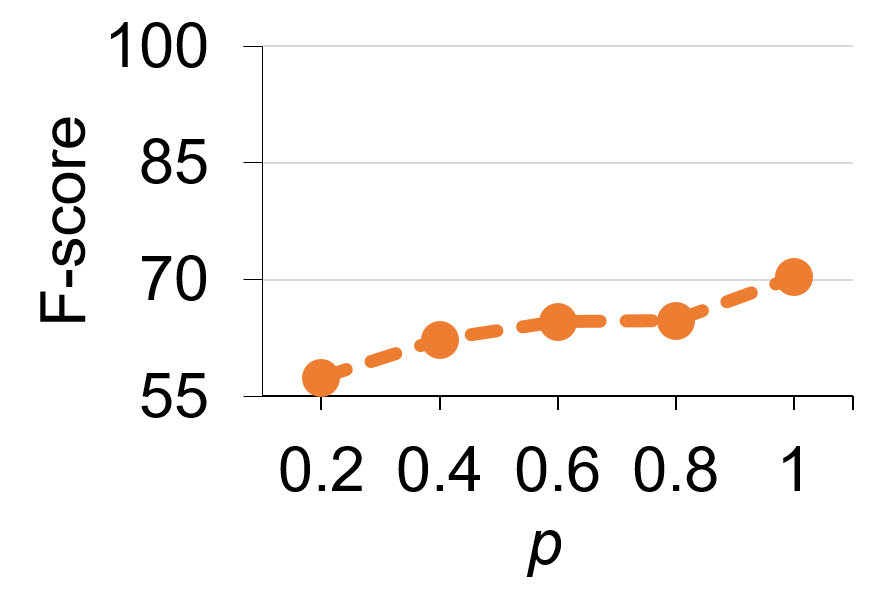}
  }
  \subfloat[GOTOV dataset]{
  \includegraphics[width = 3.6cm] {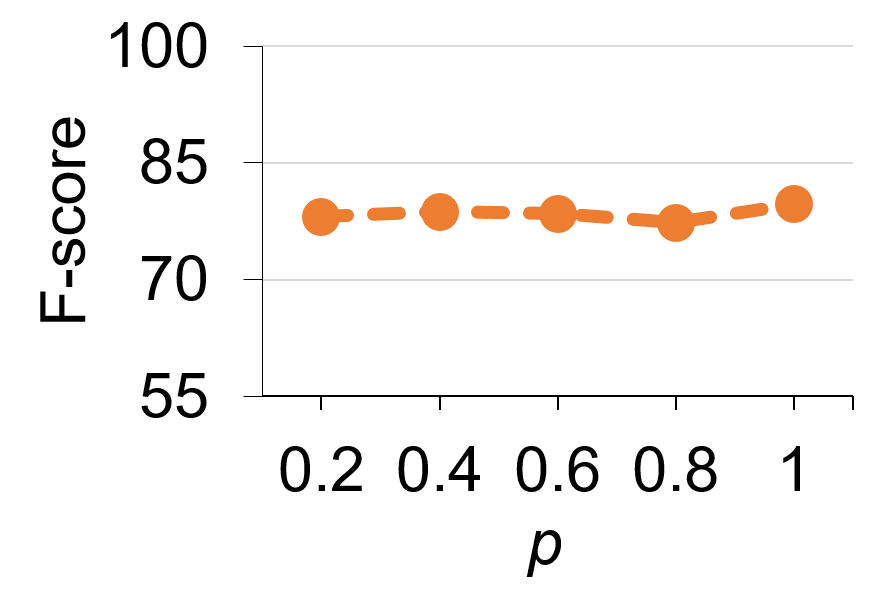}
  }
  \subfloat[Skoda dataset]{
  \includegraphics[width = 3.6cm] {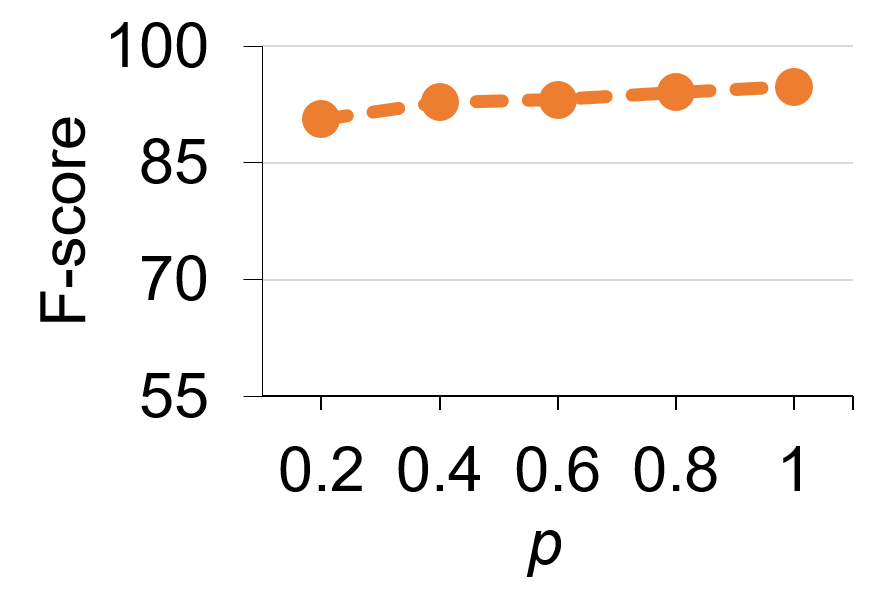}
  }
  \subfloat[Opportunity dataset]{
  \includegraphics[width = 3.6cm] {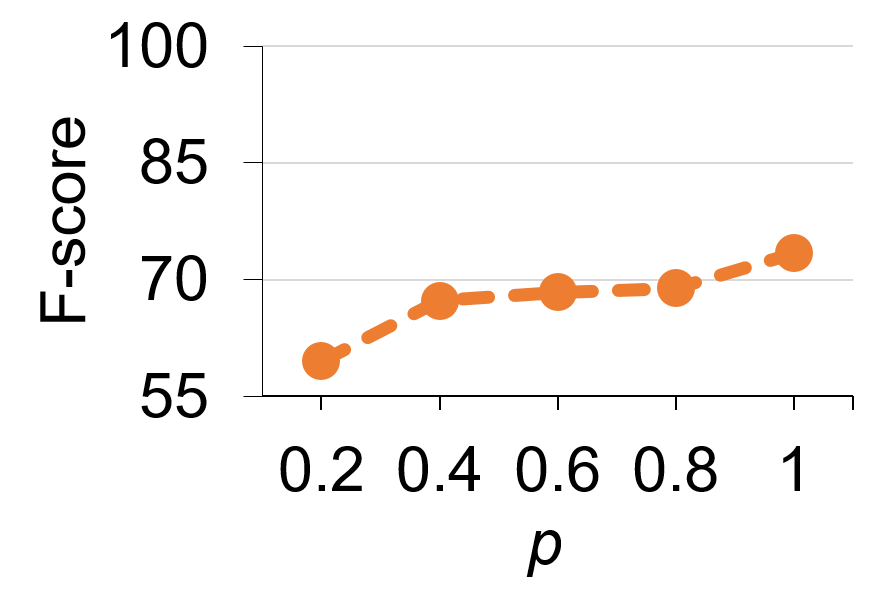}
  }
\caption{
The performance analysis of the proposed CIE-enhanced framework on limited training  data across different datasets. \label{fig:lim-EXT}}

\end{figure}

To show the effectiveness of the model on the limited amount of training data, we also investigate the performance of our proposed simpler framework ALAE-TAE-CutMix\textsuperscript{+} and CIE-enhanced framework ALAE-CIE-TAE-CutMix\textsuperscript{+} on a randomly selected $p$ amount of the training data ($p \in (0.2, 0.4, 0.8, 1.0)$) across all the considered datasets. The results of  simpler framework are presented in Figure~\ref{fig:lim-ALAE} and CIE-enhanced framework in Figure~\ref{fig:lim-EXT}. 
As HAR datasets are highly imbalanced, we ensure that all classes must have participation in the training data when evaluating the model on different data sizes. 

It is a general perspective that whenever the model is trained on a smaller amount of data, its performance is negatively impacted. However, surprisingly, the results across the GOTOV and SKODA datasets demonstrate that even with highly limited training data (only $20\%$ of the data), we observe no significant negative impact on the models' performance compared to utilizing $100\%$ of the training data$-$highlighting the effectiveness of our models. On the other hand, when examining the Hospital and Opportunity datasets, we observe some decrement in the models' performance. The most probable reason is that these two datasets are immensely imbalanced. For instance, as the reader can see in Figure~\ref{fig:data}, in the case of the Opportunity dataset, more than $75\%$ of the total data belongs to the Null class, and the rest of the remaining data belongs to the other 17 classes, which is only less than $25\%$ of the total training data. In the case of the Hospital dataset, around $80\%$ of the total data belong to two classes (setting and walking activity), and only the remaining $20\%$ of the data belongs to the other $5$ classes. Under this imbalance dataset situation, when the training data is also reduced ({\it e.g.,} $20\%$), the representation of minority classes will almost vanish. As a result, the model does not get sufficient training on the minority class data. Nevertheless, our proposed HAR models still provide acceptable performance. 

\subsection{Impact of Different Segment Sizes on Model Performance}
 
\begin{figure}[t]
\centering
\subfloat[Hospital dataset]{
  \includegraphics[width = 4.96cm] {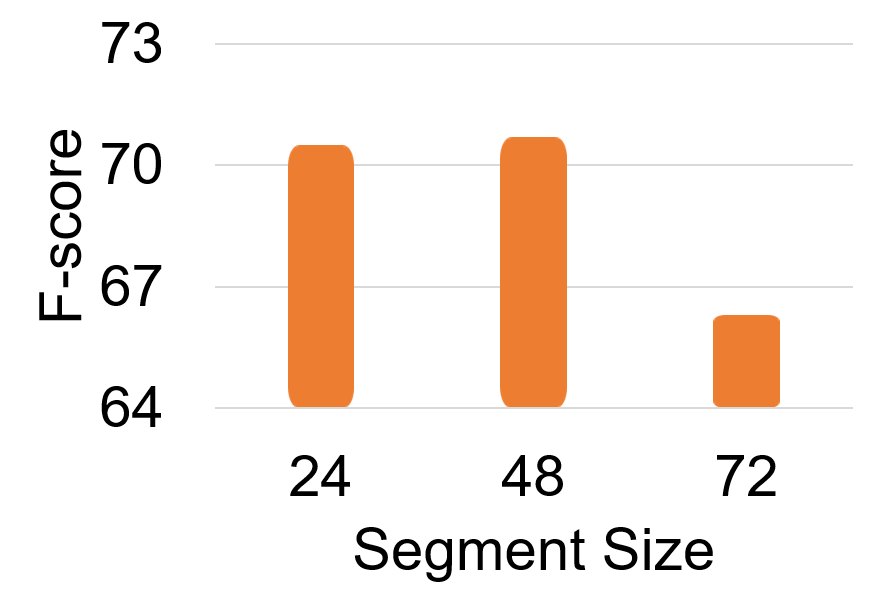}
  }
  \subfloat[GOTOV dataset]{
  \includegraphics[width = 4.96cm] {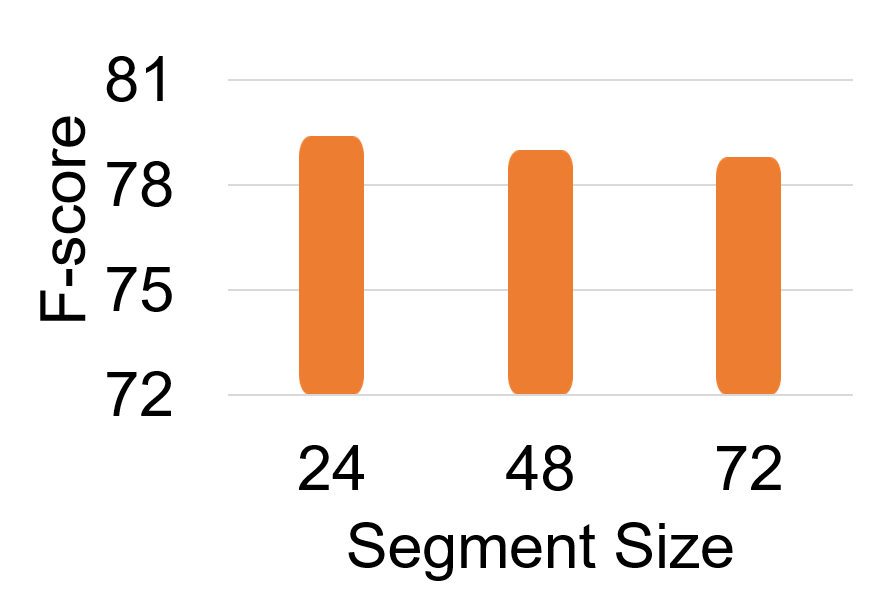}
  }
\caption{
Impact of different segment sizes on the performance of the proposed simpler framework. \label{fig:ws-alae}}
\end{figure}

\begin{figure}[t]
\centering
  \subfloat[Hospital dataset]{
  \includegraphics[width = 4.96cm] {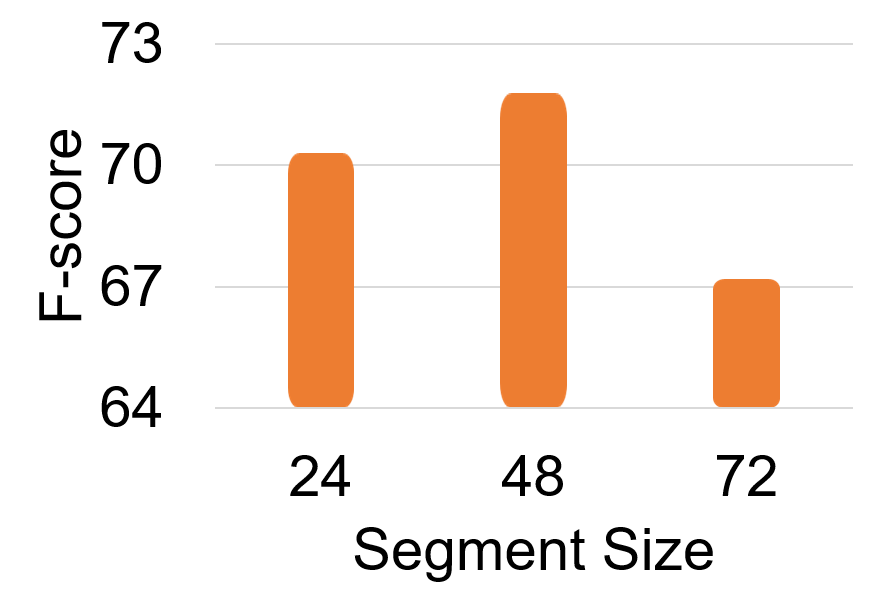}
A  }
  \subfloat[GOTOV dataset]{
  \includegraphics[width = 4.96cm] {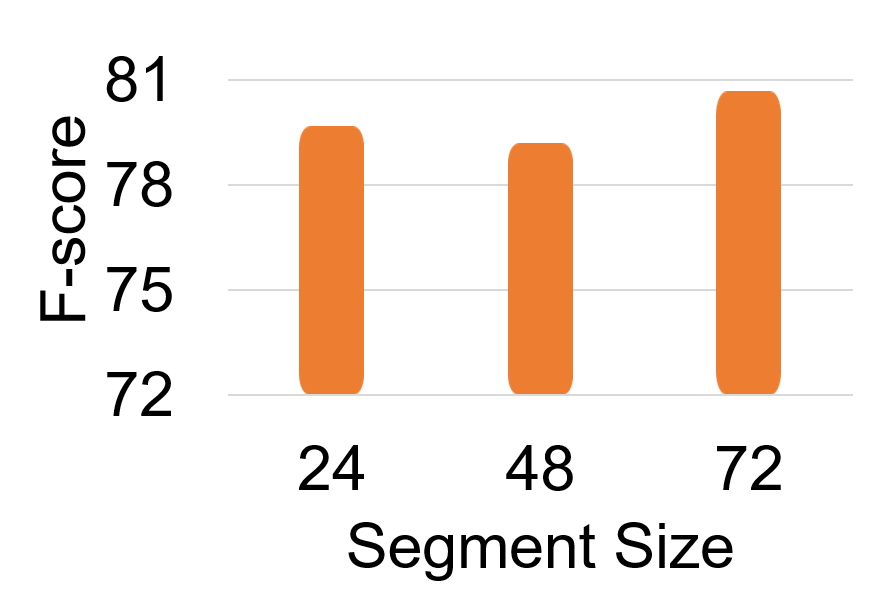}
  }
\caption{
Impact of different segment sizes on the performance of the proposed CIE-enhanced framework. \label{fig:ws-ex}}

\end{figure}

To determine the effect of different segment sizes on model performance, we also conduct experiments in which the input segment sample size is increased to $24$, $48$, and $72$ samples, respectively. The results of the proposed simpler framework are presented in the Figure~\ref{fig:ws-alae} and the CIE-enhanced framework in Figure~\ref{fig:ws-ex}.   In the case of the Hospital dataset, the results indicate that the models perform better when increasing segment size from $24$ to $48$. The reason can be that when samples are increased, the models are more confident to classify the patterns of similar nature activities, resulting more  performance gain. This observation can be validated from the results when the samples are few ($24$ samples) in a segment, where the models' performance is low compared to what is achieved on segment size $48$. 
It is due to the models confusing the activity with its similar nature activity ( see Figures~\ref{fig:con} and ~\ref{fig:E_con} {\it e.g.}, sitting with sitting down and lying with lying down activities). On the other hand, when increasing the samples too much ({\it i.e.}, $24$ to $72$), the models performance degrades. It is understandable, as the Hospital dataset includes activities whose representations are limited in the dataset and these activities are also characterized by short gestures. In this situation, increasing the samples too much in segments causes multi-class segments problem  (model receiving segments containing samples from different classes)~\cite{varamin2018deep,sdense2018}.

Surprisingly, on the GOTOV dataset, we observe better performance, particularly when increasing  samples in segments (here $72$ samples) on a CIE-enhanced model. Notably, the GOTOV dataset contains relatively more classes with characteristics of inter-class similarity problems (see Figure~\ref{fig:data}); therefore, adopting more sample  segments results the models more accurately recognizing these activities.
On the other hand, the models does not face the same multi-class window problem on this dataset that it faces on the Hospital dataset over $72$ sample segments because it is well-balanced (see Figure~\ref{fig:data}), with mostly classes that have an equal distribution and contain proper activities, not gestures.

\section{Conclusion} \label{sec:CON}
We propose a HAR framework called ALAE-TAE-CutMix\textsuperscript{+}. The ALAE module enriches the representation of each sensor channel by generating multiple latent representations. It exploits those sensor channel representations that are more discriminatory than others for the current undergoing activity in order to classify activities more precisely. The TAE module learns temporal contextual information. CutMix\textsuperscript{+} is a new augmentation method for multi-sensor channels based HAR to achieve better generalization with two variants: CutMix and CutMix\textsuperscript{+}.  CutMix is for HAR model regularization, while CutMix\textsuperscript{+} is an extension of CutMix that addresses the ambiguity in CutMix augmented sequences in some scenarios. Furthermore, we also propose the further enhanced framework, ALAE-CIE-TAE-CutMix\textsuperscript{+}, in which the ALAE-TAE-CutMix\textsuperscript{+} framework is extended by integrating the CIE module.  With the new CIE module, this framework further considers the  interactions between each pair of sensor channels. Extensive experimental results across four benchmark datasets show the remarkable superiority of both proposed frameworks' performance in effectiveness, reliability and robustness. Moreover, we also demonstrate the contribution of each module in the proposed frameworks through extensive ablation experiments. In addition, we analyze the effectiveness of the proposed frameworks in different situations by training the models with limited training data and different segment sizes.

\bibliographystyle{ACM-Reference-Format}
\bibliography{main}




\end{document}